\documentclass{article}
\usepackage{ijcai21}
\usepackage{natbib} 

\usepackage{times}
\usepackage{soul}
\usepackage{url}
\usepackage[hidelinks]{hyperref}
\usepackage[utf8]{inputenc}
\usepackage[small]{caption}
\usepackage{graphicx}
\usepackage{amsmath}
\usepackage{amsthm}
\usepackage{booktabs}
\usepackage{algorithm}
\usepackage{algorithmic}
\usepackage{tikz}
\usepackage{amssymb}
\usepackage{subcaption}
\usepackage{verbatim}
\usepackage{booktabs}
\usepackage{lastpage}
\usepackage{cancel}

\usepackage{wasysym}

\newtheorem{definition}{Definition}

\newcommand{\stacklabel}[1]{\stackrel{\smash{\scriptscriptstyle \mathrm{#1}}}}
\newcommand{\defeq}{\stacklabel{def}=}

\newcommand{\sparen}[1]{\left[ #1 \right]}

\newcommand{\set}[1]{\left\{ #1 \right\}}

\newcommand{\setsize}[1]{\left|#1\right|}

\newcommand{\given}[0]{\; | \;}

\def\final{} 

\ifdefined\final
\else
\def\enablecomments{}
\fi

\ifdefined\draft
\def\enablecomments{}
\fi

\usepackage{soul}
\usepackage[normalem]{ulem}
\usepackage{xcolor}

\definecolor{LightGreen}{rgb}{0.80,1.00,0.80}
\definecolor{LightBlue}{rgb}{0.80,0.80,1.00}
\definecolor{LightRed}{rgb}{1.00,0.80,0.80}
\definecolor{LightPurple}{rgb}{1.00,0.80,1.00}
\definecolor{LightGray}{rgb}{0.90,0.90,0.90}
\definecolor{DarkBlue}{rgb}{0.60,0.90,0.10}

\soulregister{\method}{7}
\soulregister{\xspace}{7}
\soulregister{\emph}{7}
\soulregister{\cite}{7}
\soulregister{\ref}{7}

\ifdefined\enablecomments
  \DeclareRobustCommand{\commentformat}[3]{\sethlcolor{#2}\textsf{\hl{#1: #3}}}
  \newcommand{\sm}    [1]{{\scriptsize\sethlcolor{LightGray}\hl{\textsf{#1}}}}
\else
  \newcommand{\commentformat}[3]{}
  \newcommand{\sm}    [1]{}
\fi

\newcommand{\pxm}   [1]{\commentformat{PM}{LightGreen}{#1}}

\newcommand{\zifan} [1]{\commentformat{ZW}{LightBlue}{#1}}

\newcommand{\carlee} [1]{\commentformat{CJ}{DarkBlue}{#1}}
\newcommand{\todo} [1]{\commentformat{TODO}{red}{#1}}

\title{Reconstructing Actions To Explain Deep Reinforcement Learning}

\author{
Xuan Chen\footnote{Equal Contribution}\and
Zifan Wang$^{*}$\and
Yucai Fan\and
Bonan Jin\\
Piotr Mardziel\and
Carlee Joe-Wong\And
Anupam Datta
\affiliations
Carnegie Mellon University\\
\emails
xuanche2@andrew.cmu.edu

}

\begin{document}
\maketitle
		    		    
\begin{abstract}
Feature attribution
has been a foundational building block for explaining the input feature importance in supervised learning with Deep Neural Network (DNNs), but face new challenges when applied to deep Reinforcement Learning (RL).We propose a new approach to explaining deep RL actions by defining a class of \emph{action reconstruction} functions that mimic the behavior of a network in deep RL. This approach allows us to answer more complex explainability questions than direct application of DNN attribution methods, which we adapt to \emph{behavior-level attributions} in building our action reconstructions. It also allows us to define \emph{agreement}, a metric for quantitatively evaluating the explainability of our methods. Our experiments on a variety of Atari games suggest that perturbation-based attribution methods are significantly more suitable in reconstructing actions to explain the deep RL agent than alternative attribution methods, and show greater \emph{agreement} than existing explainability work utilizing attention. We further show that action reconstruction allows us to demonstrate how a deep agent learns to play Pac-Man game.
\end{abstract}

\section{Introduction}
As machine learning algorithms become ever more ubiquitous, understanding their decisions has become critical to successful learning deployments. As a result, an increasing number of works in machine learning have developed \emph{attribution methods} that attempt to explain learning models' decisions by attributing their outputs to input features. Generally, these works focus on deep neural networks (DNNs) applied to supervised classification tasks. 
More recently, some works have attempted to apply attribution methods to deep RL (reinforcement learning) models that use a DNN to specify the optimal action an agent should take given the state of the environment~\citep{ancona2017betterv1,Bhatt2020EvaluatingAA}. The goal of such deep RL algorithms is to maximize a combination of the current and future rewards. Explaining such models is useful for applications like Atari games or autonomous driving, in which deep RL algorithms are used to determine actions taken in the game or on the road. Attribution methods can explain why the chosen action is recommended, e.g., saliency methods aim to find elements of the state that most influence the deep RL's choice of action. 
However, the evaluation of most saliency-based deep RL explanations relies on subjective judgements on whether the identified features ``should'' be considered in choosing an action~\citep{Atrey2020ExploratoryNE}. We aim to address this evaluation challenge by defining a quantitative metric to assess explanations of deep RL algorithms.


In defining an explainability metric, we further argue that directly applying DNN attribution methods to deep RL algorithms is only useful for answering questions about the specific action chosen. They do not explain \emph{why this action was chosen instead of other candidate actions}, which is particularly of interest in deep RL applications. Deep RL agents often utilize randomized policies that draw an action from a probability distribution over all possible actions, so explaining deep RL algorithms requires explaining the full behavior of the agent, not just the action taken. We define the first method to consider \emph{action reconstruction} explanations of deep RL agents by adapting DNN attribution methods to \emph{behavior-level attributions}. We then formalize an \emph{agreement} metric that quantifies how well an action reconstruction function explains the deep agent's behavior. Our method leverages the fact that deep RL agents must account for both the immediate and expected future reward of taking an action, and we show that it allows us to see how deep RL agents learn to optimize both type of rewards.

Our \textbf{contributions} include: 1) A novel class of  \emph{action reconstruction} functions that explain why a given action in deep RL was chosen over others, instead of focusing only on the specific action chosen; 2) empirical evaluations of our proposed methods on Atari games that show perturbation-based \emph{action reconstruction} outperforms existing work using attention network architectures; and 3) a demonstration that our proposed explanations and metric allow us to understand how the deep RL agent learns the optimal actions. 

\newcommand{\xvec}{\mathbf{x}}
\newcommand{\yvec}{\mathbf{y}}
\newcommand{\zvec}{\mathbf{z}}
\newcommand{\svec}{\mathbf{s}}
\newcommand{\avec}{\mathbf{a}}
\newcommand{\pivec}{\boldsymbol{\pi}}
\section{Background and Related Work}
\label{sec: background}
In this section, we give an overview of reinforcement learning algorithms and survey existing work on attribution methods, which we will later contrast with our proposed explanation methods. Throughout the paper, we use lowercase $x$ to denote scalar values and its bold font $\xvec$ to indicate vectors.

Consider an agent that interacts with the environment over a series of discrete timesteps. At each timestep $t$, the agent arrives at state $\svec_t \in \mathbb{R}^m$ and takes an action $a_t$ according to its policy $\pi(\svec_t)$, receiving a reward $r_t(a_t, \svec_t)$. Generally policies are probabilistic, and we use $ \pi(a\given\svec)$ to denote the probability of action $ a $ at state $ \svec $ according to policy $ \pi $. Our goal is to learn a policy $\pi$ that maximizes the expected total reward $\mathbb{E}\left[\sum_t \gamma^t r_t\right]$, where $\gamma\in(0,1]$ is a discount factor. We use $\mathcal{S}$ and $\mathcal{A}$ to denote the state space and action space that contain all possible states and actions, respectively. We denote the cardinality of $\mathcal{A}$, which we assume is finite, as $|\mathcal{A}|$.

\subsection{Deep Q-Network}
Two common RL approaches are policy-based approaches that learn the policy $\pi$~\citep{NIPS1999_1713,10.5555/3044805.3044850} and value-based approaches, e.g. Deep Q-Network (DQN)~\citep{mnih2013playing,Hasselt2016DeepRL}, that model the action value, which we define as the $Q$ value below:

\begin{definition}[Q value]
	\label{def: Q value}
	Given the state $\svec_t$ and rewards $r$, for an action $a_t$ under a policy $\pi$, the action value (Q value) $Q^\pi(a_t, \svec_t)$ is defined as $Q^\pi(a_t, \svec_t) \defeq \mathbb{E}\sparen{\sum^T_{\tau=t}\gamma^{\tau-t}r_\tau \given a_t, \svec_t, \pi}$
	where $T$ is the number of timesteps before the agent reaches the termination condition and $\gamma$ is the discount factor that trades off the immediate ($\tau=t$) and future ($\tau>t$) rewards.
\end{definition}

DQN is widely applied in many RL tasks~\citep{Silver2016MasteringTG,Lample2017PlayingFG,amarjyoti2017deep,Maicas2017DeepRL} due to its ability to learn complex state representations. A DQN is a network $f(\svec_t)$ that takes a state $\svec_t$ and outputs the $Q$ value for each action; we use $Q=f(\svec)$ to denote the output of the DQN for all actions $a\in\mathcal{A}$. \carlee{added this to avoid confusion later. is $\avec = \mathcal{A}$?}\zifan{No. the only place we use $\avec$ is $Q(\avec, \svec)$ which is the distribution of Q value. We use $\avec$ to emphasize $Q(\avec, \svec)$ is a distribution while $Q(a, \svec)$ is a scalar. Might have better way to say this.} A standard DQN agent follows the policy of taking the action that maximizes the $Q$ value at each step~\citep{mnih2013playing} such that $	Q(a_t, \svec_t)  \defeq \mathbb{E}\left[r_t + \gamma \max_{a_{t+1}}Q(a_{t+1}, \svec_{t+1})\right] \label{eq: Q-value}$ where $\pi^*(a_t \given \svec_t) \defeq \mathbb{I}\left[Q(a_t, \svec_t) = \max_{a_t} Q(a_t, \svec_t)\right] \label{eq: greedy policy}$.
%
%
%
%
We study DQN-based RL methods in the rest of the paper and omit the notation $t$ in $a_t, \svec_t$ for simplicity if not further noted.

\subsection{Attributions}
\todo{Do not introduce QoI}
\sm{Input attribution is a class of explanations.}
One approach for explaining the output of a DNN in supervised learning is to attribute the pre- or post-softmax output of a particular class over each  input feature. We consider the following types of attribution methods.

\noindent\textbf{Gradient-based Attribution.} 
\textit{Distributional influence}~\citep{Leino2018InfluenceDirectedEF} provides a unified view of gradient-based attributions.

\begin{definition}[Distributional Influence]
\label{def: DI}
Given a continuous and differentiable function $h(\xvec)$, and an user-defined distribution of interest $\mathcal{D}_\xvec$ related to $\xvec$, the Distributional Influence is defined as $g(\xvec, h) \defeq \mathbb{E}_{z\in \mathcal{D}_\xvec} \frac{\partial h(\zvec)}{\partial \zvec}$.
\end{definition}
For a classifier $F(\xvec)=\arg\max_c h_c(\xvec)$, the pre- or post-softmax scores of the predicted class $j$ are frequently used as $h$. For a network predicting the action of an agent, we use the pre- or post-softmax Q-value of the action of interest $f_a(\svec)$ as $h$. We can then define the following attribution methods, using subscripts to refer to a specific choice of $\mathcal{D}_\xvec$:
\begin{itemize}
    \item $g_{\text{S}}(\xvec, h)$: a set only containing $\xvec$, where Def.~(\ref{def: DI}) reduces to Saliency Map (SM)~\citep{1250762,simonyan2013deep}. One practice of applying SM in deep RL is done by \cite{wang2015dueling}.
    \item $g_{\text{IG}}(\xvec, h)$: a set containing points uniformly distributed on a linear path from an baseline input $\xvec_b$ to $\xvec$, where Def.~(\ref{def: DI}) reduces to Integrated Gradient (IG) when taking the element-wise product between $(\xvec-\xvec_b)$ and $g(\xvec)$  ~\citep{sundararajan2017axiomatic}.
    \item $g_{\text{SG}}(\xvec, h)$: a set containing points following a Gaussian Distribution $\mathcal{N}(\xvec, \Sigma)$, where $\Sigma$ is chosen by the user and Def.~(\ref{def: DI}) reduces to Smooth Gradient (SG)~\citep{smilkov2017smoothgrad}. 
\end{itemize}

\noindent\textbf{Perturbation-based Attribution.} Another line of work defines feature importance by creating counterfactual input with a perturbation set of interest and comparing the change of the model's outputs for the original and perturbed inputs. In this work we are interested in the following methods:
\begin{definition}[Occlusion-N (OC-N)]
	\label{def: occlusion-1}\citep{Zeiler2014VisualizingAU}
	Given a general function $h(\xvec)$, a user-defined baseline input feature value $b$, an integer $N$ and a partition of input features $\mathcal{E} $ with $|E| = N$ for every $ E \in \mathcal{E} $, the \emph{Occlusion-N} for $\mathcal{E}$ is defined as $	g_\emph{OC}^{\mathcal{E}}(\xvec, h) \defeq \frac{1}{N} \sum_{E\in\mathcal{E}} \mathbf{1}_E \left[\phi(h(\xvec)) - \phi(h(\xvec_{-E}))\right]$
	where $\xvec_{-E}$ is a counterfactual state by perturbing $x_i$ with $ b $ in $\xvec$ for all $i \in E$, and $ \mathbf{1}_E $ is a vector if size $ m $ equal to $ 1 $ for indices corresponding to features in $ E $ and $ 0 $ for the other indices.


\end{definition}
OC-1 is the most common choice, where we perturb each feature individually and treat the output difference as the attribution score for each feature. For the baseline value, common choices are zeros or random noise. We will use OC-1 in the experiment section. We can use the pre- or post-softmax Q-value of the predicted action $\max_a f_a(\svec)$ as $h$ and write $g_\emph{OC}(\xvec)$ to denote OC-1 for simplicity. 

We also include SARFA~\citep{Puri2020ExplainYM}, a recently proposed work that specifically targets deep RL. For OC-1 and SARFA, when the input state is an image, recent works propose to use \emph{template matching}~\citep{10.5555/1643435} to find a subset of pixels that is semantically meaningful to perturb~\citep{Iyer2018TransparencyAE} and use Gaussian blurs~\citep{Greydanus2018VisualizingAU} to create the counterfactual states. We specify a set of special treatments we employ in our evaluations in Sec.~\ref{sec: experiments}.


\begin{definition}[SARFA]~\citep{Puri2020ExplainYM}
\label{def: SARFA}
Given a Q-network $f(\svec)$, a user-defined baseline state feature value $b$ and an action $a$ of interest, consider $\sigma(\cdot)$ as a softmax function and $\sigma_{-a}(\cdot)$ as a softmax function excluding $f_a(\svec)$, SARFA $g_{\emph{SA}}(\svec)$ is defined as $g_{\emph{SA}}(\svec, f_a) = \frac{2K\Delta p}{K + \Delta p}$ where $\Delta p = \sigma(f(\svec))_a - \sigma(f(\svec_{-i}))_a$, $	K = 1 / [1 + D_{KL}(\sigma_{-a}((f(\svec)))]$ and $\svec_{-i}$ is a counterfactual state by perturbing $x_i$ with $ b $.
	                         
\end{definition}

We choose these attribution methods since they are implementation-invariant to any DQN architecture. Additionally, IG and SG were proven to outperform SM along some desirable criteria~\citep{Yeh2019OnT,wang2020smoothed}. These methods are also widely available in many open-source libraries, e.g. Captum~\citep{captum2019github} and TruLens~\citep{Trulens}, and many explainable DQN works, as cited above, employ instantiations of these methods with minor variations. 
\section{Interpreting Deep RL}
\label{sec:back-evaluation}
In this section, we firstly discuss the limitations of solely using attributions to explain a deep RL agent's behavior. We then propose a new approach, \emph{action reconstruction}, to interpret the action of a Deep RL agent based on existing attribution methods. We begin by describing criteria that motivate our approach. We further describe two action reconstruction functions that satisfy these criteria. We end this section by discussing how to evaluate our proposed approach.

\subsection{Beyond Feature Importance}
\label{sec: beyond feature importance}
\begin{figure}[t]
	\centering
	\begin{subfigure}{.22\textwidth}
		\centering
		\includegraphics[scale=0.17]{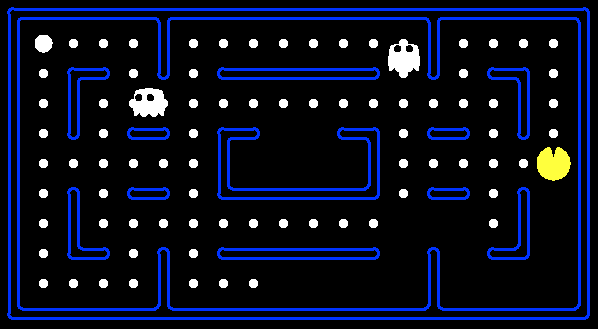}
		\caption{original input (go\_west)}
	\end{subfigure}
	\begin{subfigure}{.22\textwidth}
		\centering
		\includegraphics[scale=0.17]{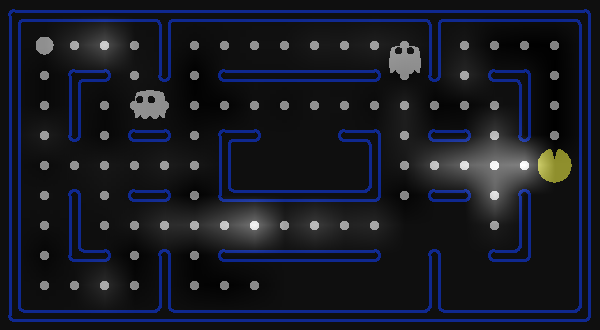}
		\caption{OC-1}
	\end{subfigure}

	\begin{subfigure}{.22\textwidth}
		\centering
		\includegraphics[scale=0.17]{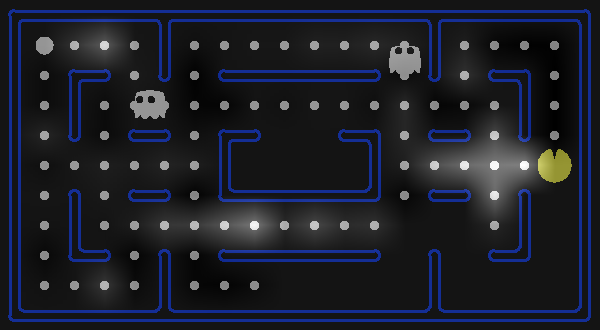}
		\caption{SARFA}
	\end{subfigure}
	\begin{subfigure}{.22\textwidth}
		\centering
		\includegraphics[scale=0.17]{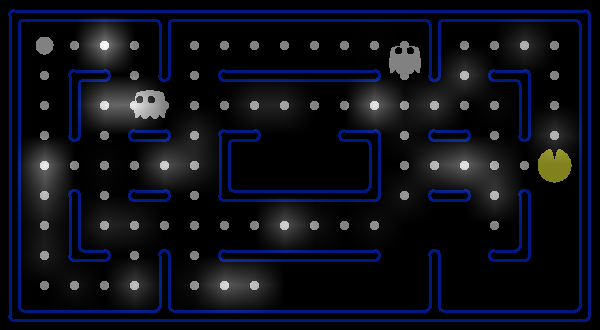}
		\caption{IG}
	\end{subfigure}

	\begin{subfigure}{.22\textwidth}
		\centering
		\includegraphics[scale=0.17]{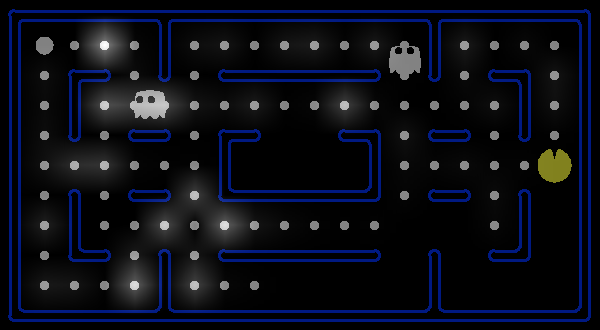}
		\caption{SG}
	\end{subfigure}
	\begin{subfigure}{.22\textwidth}
		\centering
		\includegraphics[scale=0.17]{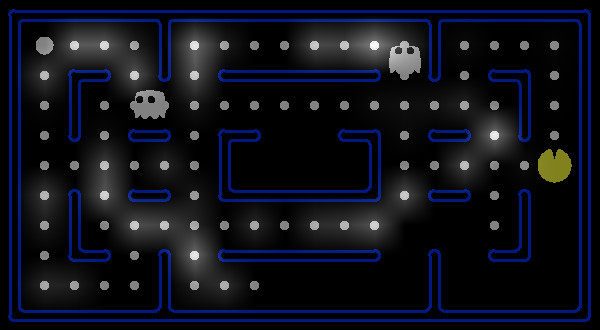}
		\caption{SM}
	\end{subfigure}
	\caption{Visualization of feature attribution maps. Important features are highlighted by
          attribution scores toward \texttt{go\_west} action using different attribution methods.
          Brightness is proportional to the attribution score.}
	\label{fig: action-level}
\end{figure}

The most popular way of interpreting attributions to human users in explaining a classification result is to highlight input features with high attribution scores. A similar approach has been adapted for analyzing important state features in deep RL, e.g. Fig.~\ref{fig: action-level}. We notice that some questions can be answered
by localizing features with high contributions, e.g.: \textit{what are the most important features that drive the
  agent to take the current action?}
On the other hand, more sophisticated questions can not be directly answered by the localization, e.g: \textit{what will the agent do given the current state}. To answer the second question, recent work in interpreting an attribution map relies mostly on having human users explain the agent's decision by combining attribution map with their prior knowledge of the game, e.g: people may infer that \textit{a high attribution score indicates the agent is going to consume that food}. However, human interpretations may lead to disagreements among users and have been shown to be misleading with counterfactual experiments, particularly in deep RL games~\citep{Atrey2020ExploratoryNE}.

In this paper, we aim to resolve such questions by proposing a new approach of interpreting attributions for Deep RL agents, which we call \emph{action reconstruction}. Unlike attributing a Q-value over the state features, our approach is motivated from using proxy models to explain the network's behavior, e.g. LIME~\citep{Ribeiro2016WhySI}, where the explainer develops a simple model serving as a proxy to the target model in the neighborhood of the given input. In the context of Deep RL, we look for an action reconstruction function $\hat{f}: \mathcal{S} \rightarrow \mathcal{A} $ that serves as a proxy to the target DQN $f: \mathcal{S} \rightarrow \mathcal{A}$. LIME is one possible choice, but the following concerns prevent us from applying LIME to deep RL: it is not computationally efficient compared to the feature attributions mentioned in Sec.~\ref{sec: background}, and random perturbations around the input may capture the DQN's behavior out of the data manifold, i.e., those perturbed states may not be valid states in the environment. Recent work demonstrates that different attribution functions $g_*$ can be viewed as finding a linear model with a particular perturbation set of inputs to model the output score (e.g. Q-value of each action) with a certain degree of faithfulness~\citep{Yeh2019OnT}. We therefore begin with a feature attribution $g_*$ and extend it to model the actual action to complete the construction of $\hat{f}: \mathcal{S} \rightarrow \mathcal{A}$.

\subsection{Criteria of Action Reconstruction}\label{sec:methods}

We consider the following criteria as natural requirements for an action reconstruction model $\hat{f}$:
\begin{itemize}
    \item \textbf{User-Invariance.} The output of the shadow model $\hat{f}$ is invariant to the human user who makes a query for the explainability. This criterion rules out inferring $\hat{f}$ from a visualization of the attribution map as discussed in Sec.~\ref{sec: beyond feature importance}.
    
    \item \textbf{Implementation-Invariance.} The construction of $\hat{f}$ is invariant to the network architecture and training routines applied to $f$. This criterion guarantees the generality of the approach.
    \item \textbf{Proportional to Reachability.} Suppose a well-defined \emph{reachability} $r(\svec)$ (Def.~\ref{def: intractability}) measures the agent's ability to interact with state features at the current time. Then the output of $\hat{f}$ depends more on features with higher \emph{reachability} scores.
\end{itemize}

\begin{definition}[Reachability]
\label{def: intractability}
  Reachability is a mapping $r:\mathcal{S} \rightarrow \mathbb{R}^{|\mathcal{S}|}_{+}$. That is, the reachability score for any state features is non-negative.
\end{definition}

\paragraph*{Why Proportionality?} The first two criteria are clear from their definitions; therefore, we focus on explaining the last criterion. In visualizing attributions for an agent playing Pac-Man, we empirically find that features, e.g., \texttt{food} or \texttt{ghost}, that are very far away from the agent can receive very high attribution scores in Fig.~\ref{fig: action-level}(b) and (e), compared to those close to the agent. Prior knowledge of the game lets us know that the Pac-Man agent needs to take a significant number of steps to actually interact with those far features, e.g., consume a \texttt{food}. Other games exhibit similar scenarios. For example, in Breakout a ball can interact with a break if and only if the ball clears blocking breaks below or above the target break. In order to be convincing to users, our action reconstructions should be consistent with this prior knowledge of the environment. 
Our reachability function $r(\svec_i)$ formalizes this intuition by measuring the ability of a given state feature that may influence the agent's decision at current step based on the prior knowledge of the environment. We show specific examples of $r(\svec_i)$ in Sec.~\ref{sec: experiments} for the games evaluated in this paper.

\subsection{Action Reconstruction}
The answer to \emph{what the agent will do given a state} can be further decomposed into a much simpler question: \emph{what the agent will do given each individual state feature}. Therefore, before we introduce the construction of $\hat{f}$, we first introduce \emph{behavior-level attribution}.
\begin{definition}[Behavior-Level Attribution]
  Given an action space $ \mathcal{A} $, state space $ \mathcal{S} $ a \emph{behavior-level
    attribution} method is a mapping $G: \mathcal{S} \rightarrow \mathcal{A}^m $.
\end{definition}


\sm{One goal of behavior-level attribution.} 
That is, given a state, the behavior-level attribution produces an action for each state feature (in the same way an attribution produces a real-numbered score for each state feature). The definition itself does not impose any restrictions on behavior-level attributions, but the
metrics which we will define over them will presume a goal: they should indicate the association
between an input feature and a specific action if an agent ``sees" that feature individually. For
example, given a Pac-Man agent, when considering each feature individually, the food on the agent's
north side likely attracts the agent to \texttt{go\_north} while a ghost at the same position would
be expected to have the association with the move in the opposite direction.

\paragraph*{Behaviors from Actions.}
A behavior-level attribution can be constructed from any attribution defined in Sec.~\ref{sec: background} in the expected
manner. One intuitive way is to find the action towards which the state features create the highest contribution, therefore, give an attribution $ g_* $, we define $G_*(\svec)_i \defeq \arg \max_a g_*(\svec, f_a)_i$.
Motivated by our desired properties of \emph{user-invariance}, \emph{implementation-invariance} and \emph{proportionality}, we now formally introduce the $\lambda$-Local Action Reconstruction function $\hat{f}_\lambda$ and a smoothed version, Kernel Action Reconstruction $\hat{f}_k$.

\begin{definition}[$\lambda$-Local Action Reconstruction] ($\lambda$-LAR)
  \label{def: Locally Reconstructed Action}
  Given a behavior-level attribution $G: \mathcal{S} \rightarrow \mathcal{A}^m $ for some state $\svec$ and the reachability function $r$ incorporating the prior knowledge of the environment,
  \emph{$\lambda$-Local behavior-level Action Reconstruction} $\hat{f}_\lambda $ is defined as $\hat{f}_\lambda(G(\svec))  \defeq
    \arg\max_a \setsize{\set{i: i \in [m], r(\svec)_i \leq \lambda, G(\svec)_i = a}}$.
\end{definition}

\begin{definition}[Kernel Action Reconstruction] (KAR)
  \label{def: Locally Reconstructed Action}
  Given a behavior-level attribution $G: \mathcal{S} \rightarrow \mathcal{A}^m $ for some state $\svec$, the reachability function $r$ incorporating the prior knowledge of the environment and a kernel funciton $k$, the
  Kernel behavior-level Action Reconstruction $\hat{f}_k $ is defined as $\hat{f}_k(G(\svec)) \defeq
    \arg\max_a \sum_j R_{a, j}$
  where $R_a \defeq \set{k(r(\svec)_i, b) \given i: i \in [m], G(\svec)_i = a}$ and $b$ is an user-defined baseline interaction ability
\end{definition}



\pxm{What is the local reconstruction action as defined here is not the best way to reconstruct an
  action given the local area?}

\paragraph*{Evaluation.}


Whereas the criteria for evaluating an attribution method $g_*$ are whether $g_*$ satisfies different desired properties, e.g. sensitivity~\citep{ancona2018towards}, fidelity~\citep{Yeh2019OnT}, robustness~\citep{wang2020smoothed}, the evaluation criterion for action reconstruction $\hat{f}$ is
to measure whether $ \hat{f}$ faithfully reconstruct the same prediction as as $f$. We therefore end this section by
introducing the \emph{agreement} score.

\begin{definition}[Agreement]
  \label{def: Alignment}
  Given a Q-network $Q(\avec, \svec)=f(\svec)$ and an action reconstruction function $\hat{f}$ built by a behavior-level attribution $G$, we define
  \emph{Agreement} $L(\hat{f})$ as $    L(\hat{f}) \defeq \mathbb{E}_{\svec \sim \mathcal{S}} \mathbb{I}\sparen{\arg\max_a Q(a,
      \svec) = \hat{f}(G(\svec))}$.
\end{definition}
Agreement as defined here has been discussed by \cite{Annasamy2019TowardsBI} in terms of the actions reconstructed by another network, e.g., an auto-encoder. Our contribution is to formally define agreement with respect to a general action reconstruction function $\hat{f}$. We use this metric to compare against \cite{Annasamy2019TowardsBI} in the next section.


\section{Experiments}
\label{sec: experiments}
In this section, we first define reachability functions for Pac-Man, SpaceInvader and Breakout~\citep{atari} to be evaluated in this section. We give empirical evidence for the Pac-Man game that the chosen reachability function incorporates our prior understanding of the game. We then show visualizations of behavior-level attribution and compare it with feature attribution maps. We thirdly search for the feature attribution $g_*$ that maximizes the agreement achieved by our action reconstructions $\hat{f}_\lambda$ and $\hat{f}_k$. We end this section by showing that our action reconstruction function can not only explain agent actions but also monitor how DQN learns to play Pac-Man.

\begin{figure}[t]
    \centering
    \includegraphics[width=0.43\textwidth]{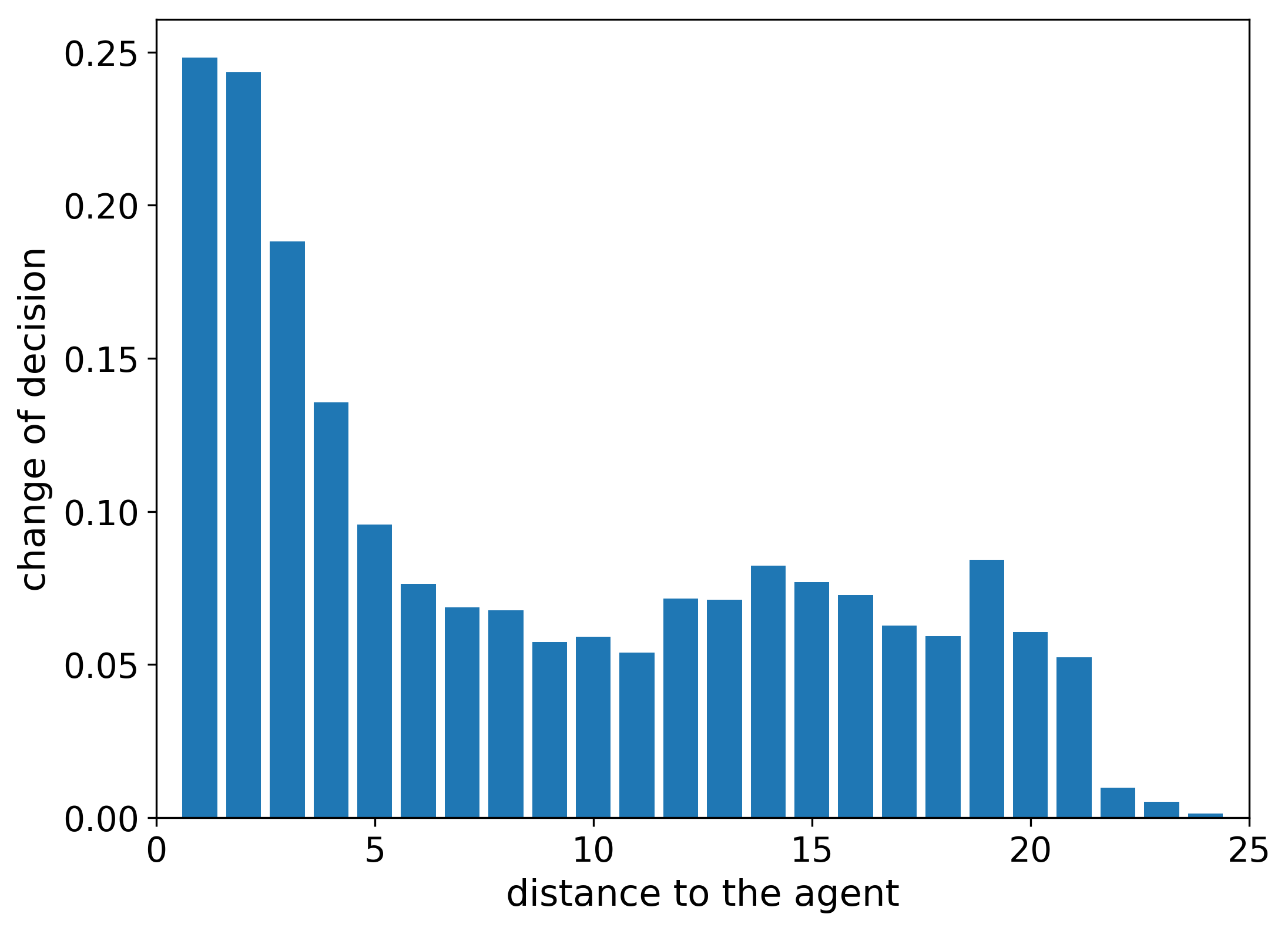}
    \caption{The percentage of predicted actions change due to the perturbation of features with equal reachability. In the Pac-Man game, we use $\ell_1$ distance from features to the agent to incorporate the reachability. }
    \label{fig: intrability}
\end{figure}

\noindent\textbf{Measurement of Reachability.}
For Atari games evaluated in this paper, we employ the following reachability measurements. For Pac-Man, since the agent is only allowed to move in the $\ell_1$ space with at most one unit length (or a constant amount of pixels in the image) per timestep, we use $r(\svec)_i = ||\mathbf{p}-\mathbf{q}_i||_1$ where $\mathbf{p}$ is the coordinate of the agent and $\mathbf{q}_i$ is the coordinate of the state feature $s_i$. For Space Invader, since the possibility of being attacked increases with a decrease of the $\ell_2$ distance, we use $r(\svec)_i = ||\mathbf{p}-\mathbf{q}_i||_2$. For Breakout, $\ell_p$ distance is not intuitively reasonable. Therefore, we use $r(\svec)_i = 1 + u(\svec)_i$ where $u(\svec)_i$ is the number  of  bricks  between  the  brick  of  interest  and  the agent, since the ball will get reflected once it hits a brick. Notice that these measurements are not unique or optimal but rather the most intuitive choices. We discuss how we may improve the design of $r(\svec)$ in Sec.~\ref{sec:conclusion}. To validate that the chosen reachability function suffices to incorporate the reachability between features and the agent, we provide the following counterfactual analysis to support our choice: For a series of states, we perturb all features with equal reachability to the agent and compare if the network's predicted action changes due to the perturbation. We include the result from the Pac-Man game as an example in Fig.~\ref{fig: intrability}. All results are aggregated over 6000 states in 20 game rounds. As illustrated in Fig.~\ref{fig: intrability}, perturbing features with weak reachability (far away from the agent in the Pac-Man game) has very limited influence on changing the agent's decision; therefore, when reconstructing actions from an attribution map, if the attribution method is not faithful enough to the agent's decision (assigning high attribution scores to features with weak reachability), relying only on attribution scores to reconstruct the action is not reliable enough. We discuss potential sources of this discrepancy between attributions and reachability in Sec.~\ref{sec: Quantitative Evaluation}.

\subsection{Visual Evaluation}
\begin{figure}[!t]
	\centering
	\begin{subfigure}{.22\textwidth}
		\centering
		\includegraphics[scale=0.17]{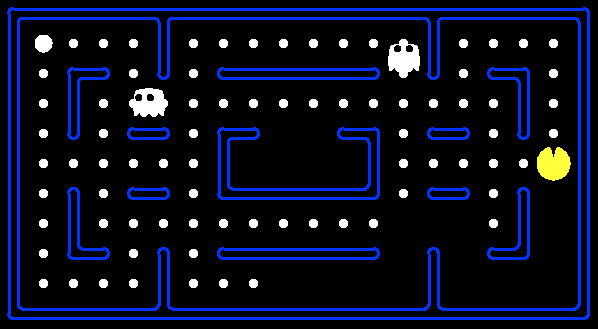}
		\caption{original input (go\_west)}
	\end{subfigure}
	\begin{subfigure}{.22\textwidth}
		\centering
		\includegraphics[scale=0.17]{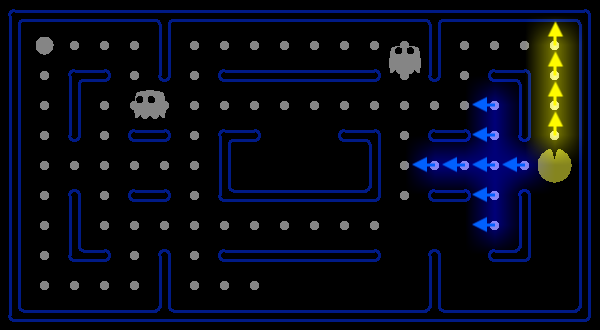}
		\caption{OC-1}
	\end{subfigure}

	\begin{subfigure}{.22\textwidth}
		\centering
		\includegraphics[scale=0.17]{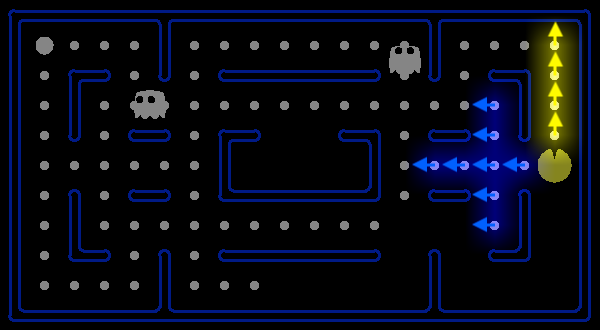}
		\caption{SARFA}
	\end{subfigure}
	\begin{subfigure}{.22\textwidth}
		\centering
		\includegraphics[scale=0.17]{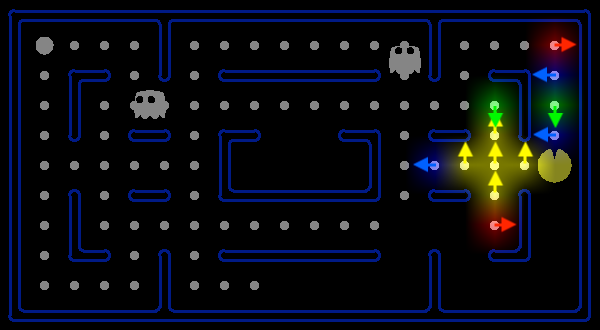}
		\caption{IG}
	\end{subfigure}

	\begin{subfigure}{.22\textwidth}
		\centering
		\includegraphics[scale=0.17]{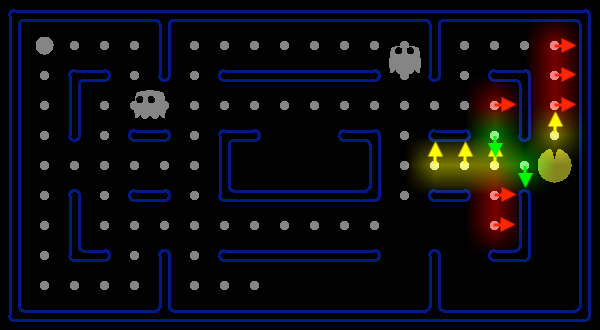}
		\caption{SG}
	\end{subfigure}
	\begin{subfigure}{.22\textwidth}
		\centering
		\includegraphics[scale=0.17]{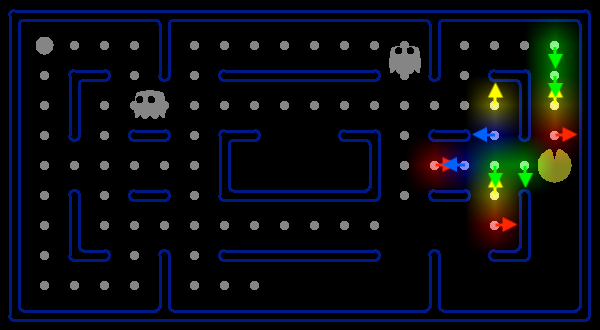}
		\caption{SM}
	\end{subfigure}
	\caption{Visualization of $\rho(\svec)$ using different attribution methods. We use colors and arrows to indicate the action each feature mostly attributes towards. \textbf{blue}: \texttt{go\_west}, \textbf{green}: \texttt{go\_south}, \textbf{red}: \texttt{go\_east}, and \textbf{yellow}: \texttt{go\_north}. We show features within $\lambda \leq 4$ by setting $r$ as $\ell_1$ distance. This figure is better visualized on screen.}
	\label{fig: behavior-level}
\end{figure}

We first demonstrate how a behavior-level attribution differs from feature attribution in the visualization. We use the attribution methods $g_*$ described in Sec.~\ref{sec: background} to build the corresponding $G_*$ in Fig.~\ref{fig: behavior-level} for the same input state used in Fig.~\ref{fig: action-level} (hyper-parameter values are given in Supplementary Material A).
Each color in Fig.~\ref{fig: behavior-level} represents an action to which this feature attributes most significantly. In this example we focus on a neighborhood around the agent with equal reachability. OC-1 and SARFA show the same results by identifying that all \texttt{food} on the west attract the agent to \texttt{go\_west} (similarly for ``north''), while SM, IG, and SG associate some \texttt{food} with an action with which the agent cannot actually consume that \texttt{food}. \textbf{Therefore, from visual examination, OC-1 and SARFA produce more reasonable behavior-level attributions}. Compared to feature attribution, behavior-level attribution labels each state feature with the action the agent is supposed to take upon ``seeing'' it. We restrict the visualization to within a neighborhood due to the $\ell_1$ distance reachability function. For other games, users should adapt the visualization according to the chosen reachability function. More visualizations are included in \nameref{E}. 

\subsection{Quantitative Evaluation}
\label{sec: Quantitative Evaluation}
In this experiment, we benchmark $\hat{f}_\lambda$ and $\hat{f}_k$ over candidate behavior-level attributions $G_*$ in terms of the agreement (Def.~\ref{def: Alignment}) on binary Pac-Man (bPM, the input game state is a set of location matrices), image Pac-Man (iPM,  the input game state is an image), Space Invader (SI) and Breakout (Br). All feature attribution scores are computed using the post-softmax output for SM, IG, SG and OC-1 (post-softmax scores uniformly outperform the pre-softmax scores and detailed comparisons are included in \nameref{D}). Results related to bPM are aggregated over 7419 frames from 30 matches and results related to iPM, SI and Br are aggregated over 6000 frames from 20 matches (the agent may experience different numbers of frames for each match). We normalize $\lambda$ in $\hat{f}_\lambda$ to [0, 1] by dividing each $\lambda$ with the maximum possible reachability in that game. For $\hat{f}_k$, we use the negative exponential kernel $k(r(\svec)_i, b) = \exp\{-[r(\svec)_i-b]^2\}$ and we set $b=0$ to produce exponential decay for features with weak reachability. Exploring other choices of kernel functions is left as future work. Hyper-parameters and implementation details are left in the \nameref{A}. We also compare the agreement scores with a baseline method where an attention structure and AutoEncoder (or VAE) are used to build a self-explainable DQN for the Pac-Man game (iPM-idqn). The agreement scores for iPM-idqn are further inferred from attention weights and AutoEncoder (or VAE)~\citep{Annasamy2019TowardsBI}. For all Pac-Man games, we only calculate the agreement scores using \texttt{food} features, which is the most common state feature in the game. We give agreement scores for other features, e.g. \texttt{ghost}, in the \nameref{B}. For other games, we use all state features. Results are shown in Table~\ref{tab: agreement}.

\setlength{\tabcolsep}{1pt}
\begin{table}[t]
    \centering
\begin{tabular}{ccccccc}
\rotatebox{45}{Game} &\rotatebox{45}{$\max_{\lambda, G} L(\hat{f}_\lambda(G))$} &\rotatebox{45}{$\lambda'$} & \rotatebox{45}{$G'$} &\rotatebox{45}{$\max_GL(\hat{f}_k(G))$} &\rotatebox{45}{$G''$ } & \rotatebox{45}{$L(\texttt{att.})$}\\
\hline\hline
bPM & .66 &.15 & OC-1 &\textbf{.68} & OC-1 & /   \\
iPM  & \textbf{.37}& .27 & OC-1 &.31 & OC-1 &/  \\
iPM-idqn & \textbf{.39}& .58  & OC-1 & .30 & OC-1 &.31  \\
SI & \textbf{.34}& .01 & SM &.31 & SG&/   \\
Br & \textbf{.46}& .60 & OC-1 &.33 & OC-1 &/ \\
\end{tabular}
    \caption{Agreement scores of $\hat{f}_\lambda$ and $\hat{f}_k$ for different games. $\lambda', G' = \arg\max_{\lambda, G}L(\hat{f}_\lambda(G))$ and $G'' = \arg\max_G L(\hat{f}_k(G))$. Full results are included in Table 4 and 5 from \nameref{C}. $L(\texttt{att.})$ denotes the highest reported agreement score from ~\cite{Annasamy2019TowardsBI}. Higher agreement scores are better.}
    \label{tab: agreement}
\end{table}

\noindent\textbf{Observations from Table~\ref{tab: agreement}.} In the $\text{2}^{\text{nd}}$ and the $\text{5}^{\text{th}}$ columns of Table~\ref{tab: agreement}, we only report the highest agreement sorted over all behavior-level attributions (and the corresponding $\lambda \in [0, 1]$, if applicable) and leave the full results in the \nameref{C}. We fill in the highest reported agreement score from ~\cite{Annasamy2019TowardsBI} under $L(\texttt{att.})$. We can then make the following observations:
1) \textbf{Focusing on the $\text{2}^{\text{nd}}$ column}: Except for SI, the highest $L(\hat{f}_\lambda)$ occurs neither close to $\lambda=0$ nor $\lambda=1$, which indicates that the well-trained agents learn to make decisions based on features within a certain "perceptional horizon" measured by reachability functions. The agent does not just rely on closest features nor on everything in the environment, and $\lambda' = \arg\max L(\hat{f}_\lambda)$ suggests ``how far an agent sees into the future". The result agrees with the training loss used for Q-learning. In SI, for example, the agent only needs to eliminate the closest enemies to survive, so the highest agreement for $\hat{f}_\lambda$ happens at a fairly small $\lambda=0.01$. 2) \textbf{Best behavior-level attribution for $\hat{f}$}: OC-1 outperforms other methods for both $\hat{f}_\lambda$ and $\hat{f}_k$ in most cases. It is the recommended method to proximate the target DQN's behavior. For bPM, the features are better-defined compared to image-based games, so the agreement scores are significantly higher than other games. 3) \textbf{Comparing $\hat{f}_\lambda$ with $\hat{f}_k$}: Finding $\hat{f}_k$ is much computationally efficient compared to $\max_\lambda \hat{f}_\lambda$, without a significant loss of agreement; however, it cannot answer precisely ``how far the agent sees into the future," which $\hat{f}_\lambda$ answers with the optimal $\lambda'$. Thus, we have a trade-off between efficient and precise explanations. 4) \textbf{Comparing with the baseline:} The $\text{3}^{\text{rd}}$ row shows that LAR performs better than reconstructions based on attention weights used by \cite{Annasamy2019TowardsBI}, while our proxy models $\hat{f}_\lambda$ are much simpler and more straightforward. Without confining ourselves  to specific architectures, we can still explain the agent's behavior in a post-hoc manner with even better agreement.

\noindent\textbf{Challenges in Distributional Influence.} $\hat{f}_\lambda$ (and $\hat{f}_k$) constructed from distributional influence do not outperform other methods in the results, which may suffer from the fact that the distribution of interest $\mathcal{D}_\xvec$ may contain a lot of inputs that are not semantically meaningful: these states may not exist in $\mathcal{S}$. On the other hand, we currently use zeros as the baseline input for IG, which is also not semantically meaningful (in contrast, zeros in image classification indicate a black image). Seeking for a meaningful baseline input can be a significant challenge in explaining reinforcement learning. 

\begin{figure}[t]
	\centering
	\includegraphics[width=0.47\textwidth]{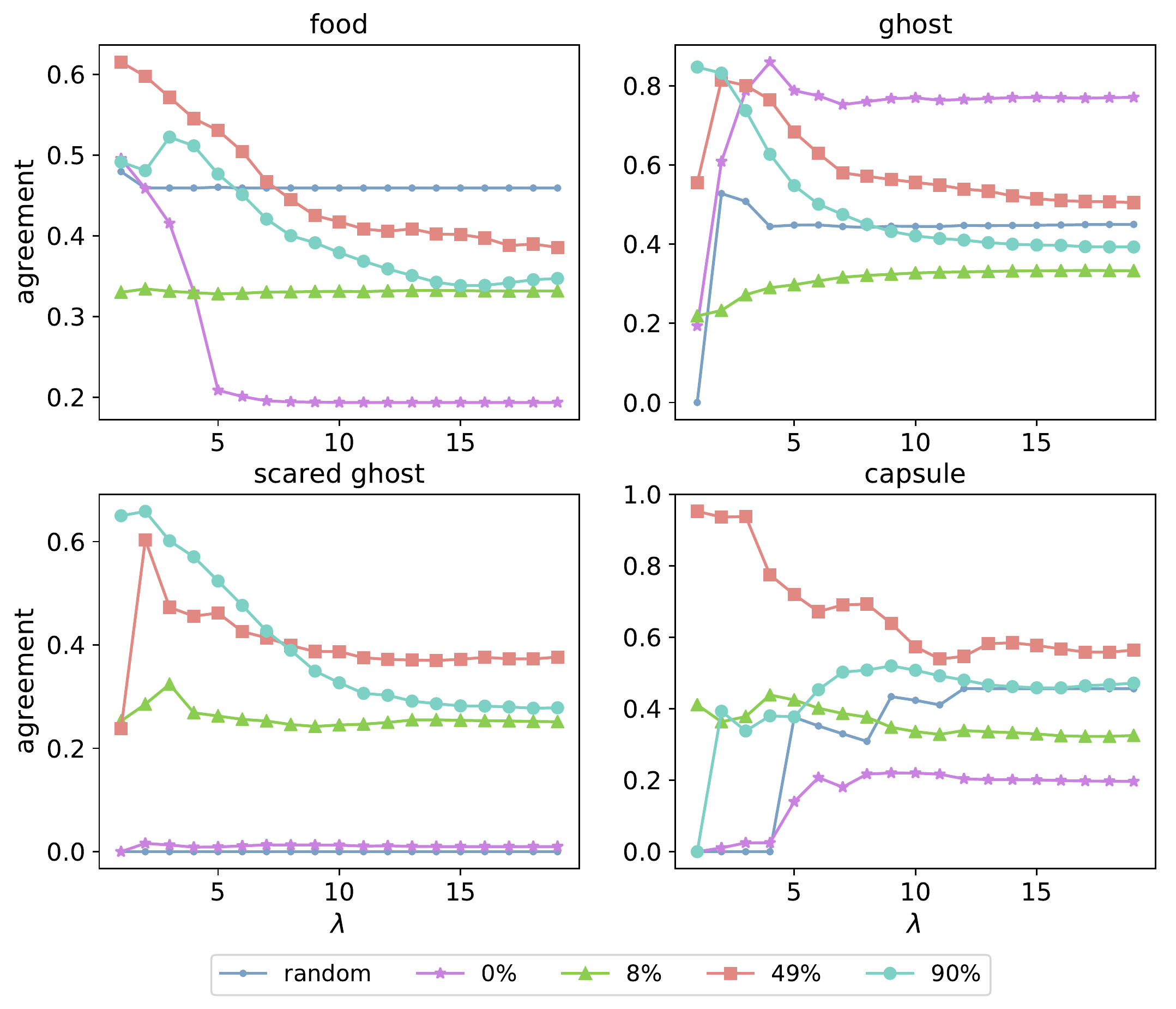}
	\caption{Agreement scores of $\hat{f}_\lambda$ of four DQNs at different training stages for four kinds of features in Binary Pac-Man game using OC-1.}
	\label{fig: training}
\end{figure}

\subsection{Beyond Explaining Action: How does DQN Play Pac-Man?}
We demonstrate how to leverage \emph{action reconstruction} to monitor the learning process of the Pac-Man agent. We trained 4 DQN models (without attention) to play bPM with different training epochs, which have success rates of 0\%, 8\%, 49\% and 90\% over 100 test matches. We also add a baseline model where the weights are randomly initialized without performing any training. We use $\hat{f}_\lambda(G_\text{OC})$ since it performs best in the previous experiments. The results are shown as plots in Fig.~\ref{fig: training}, from which we make the following observations: 1) Compared with other agents, at the beginning of the training when the success rate is 0\%, the agent learns to align its action with the position of \texttt{ghost} features instead of \texttt{food} ones, given that the agreement score on the \texttt{ghost} feature is close to other well-trained models and the curve is above \texttt{random}. In other words, the Pac-Man agent learns to survive first (by avoiding ghosts) instead of winning (by consuming food). 2) The agreement curves on \texttt{scared\_ghost} shift up when the success rate increases from 0\% to 90\%, showing that the agent gradually learns to consume  \texttt{scared\_ghost} to earn higher rewards.
\section{Conclusions}\label{sec:conclusion}
In this work, we first identify explainability questions that existing feature attribution methods are not capable of answering. Motivated by proxy models in explaining DNNs, we introduce and construct \emph{action reconstruction} functions, $\lambda$-LAR and KAR, as transparency tools to explain deep RL agents. Our empirical results show that OC-1 outperforms other methods in building such functions and outperforms baseline explanation methods. In practice, we show that by leveraging \emph{action reconstruction}, a human user can quantitatively monitor the learning process of a deep RL agent.
\section*{Acknowledgements}

This work was developed with the support of NSF grant CNS-1704845 as well as by DARPA and the Air Force Research Laboratory under agreement number FA8750-15-2-0277. The U.S. Government is authorized to reproduce and distribute reprints for Governmental purposes not withstanding any copyright notation thereon. The views, opinions, and/or findings expressed are those of the author(s) and should not be interpreted as representing the official views or policies of DARPA, the Air Force Research Laboratory, the National Science Foundation, or the U.S. Government. 

We gratefully acknowledge the support of NVIDIA Corporation with the donation of the Titan V GPU used for this research.

\section*{Ethical Impact}
Our work enables (human) users or observers of RL algorithms to better understand an RL agent's decisions, by identifying the state features that contribute to the agent behavior and quantifying the degree to which these explanations match the true RL agent behavior. These explanations can help increase users' trust in RL decisions, and our metrics for the explanation quality can further inform users of how seriously they should treat an explanation method's findings. Conversely, the results of our methods, when applied to specific realizations of deep RL agents, can be used to scrutinize the ethics of these agents' behavior, by revealing the underlying reasons behind agent decisions and the degree to which these reasons are accurate.

\bibliographystyle{named}
\bibliography{ijcai21}
\clearpage
\appendix
\section*{Technical Appendix}

\subsection*{Technical Appendix A}
\label{A}
\begin{figure}[!h]
	\centering
	\includegraphics[width=0.3\textwidth]{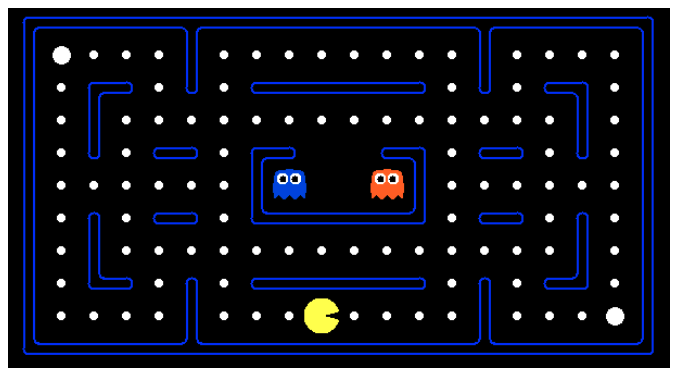}
	\caption{The layout of medium grid}
	\label{fig: example}
\end{figure}

In this section, we first introduce the binary environment we use and give the hyper-parameters and network architectures for both environments. We also report the parameters for the attribution approaches(SG and IG). For experiments on binary Pac-Man, we use environment produced by \citet{van2016deep} \footnote{\url{https://github.com/tychovdo/PacmanDQN}}, which is adopted from the project created by UC Berkeley \footnote{\url{http://ai.berkeley.edu/project_overview.html}}.
This PacMan implementation has native support for different maps or levels. Our paper utilize the medium-map (Figure \ref{fig: example}) which is a game grid of 20 $\times$ 11 tiles. The original version of PacMan has a size of 27 $\times$ 28 game grid.

\paragraph{State Representation} Each game frame consists of a grid or matrix containing all 6 features. In each matrix a 0 or 1 respectively express the existence or absence of the element on its corresponding matrix. As a consequence, each frame contains the locations of all game-elements represented in a $W \times H \times 6$ tensor, where $W$ and $H$ are the respective width and height of the game grid. Conclusively a state is represented by a tuple of two of these tensors together representing the last two frames, resulting in an input dimension of $W \times H \times 6 \times 2$.
\paragraph{Network architecture} 
The Q-network and the target network consist of two convolutional layers followed by two fully connected layers. The parameters of each layer is shown in Table \ref{tab:Binary Architecture}.

\paragraph{Training Parameters} 
For the professional agent and the one we use to compare with other attribution methods, we set $\gamma=0.95$.

The replay memory we use has a maximum memory of 10000 experience tuples to limit the memory usage. To ensure this replay memory is filled before training, the first Q-function update occurs after the first 5000 iterations. Every training iteration a mini-batch, consisting of 32 experience tuples, is sampled. 

\paragraph{Attribution Methods}
For Integrated Gradient, we use the all zero matrix as the baseline and set the step to be 50. For Smooth Gradient, the noise we add to the original input follows a Gaussian distribution centered at $\svec$ with variance $\sigma^2 = 0.15|\max(\svec) - \min(\svec)|$.

\paragraph{Atari Games} For experiments of three Atari environments-MsPacman(image Pac-Man), Space Invaders, and Breakout, we adopted the gym library \footnote{\url{https://gym.openai.com/envs/\#atari}}. The code and pre-trained RL agents we use are available at \url{https://github.com/greydanus/visualize atari}. These agents are trained using the Asynchronous Advantage Actor-Critic Algorithm (A3C). The parameters of each layer for the Q-network and the target network are shown in Table \ref{tab:Atari Architecture}. We follow their default hyperparameters during training. The average 100 episode's rewards for the agents we use to evaluate are $3323.01$, $539.30$, and $1808.45$ for MsPacman, Space Invaders, and Breakout respectively. For the gradient-based attribution methods, we use the same parameters above.

\begin{table}[t!]
  \centering
  \begin{tabular}{ccccc}
  \hline
    Type & Kernels & size & Stride\\
    \hline
    Convolutional & 16 & $3\times 3$ & 1 \\
    Convolutional & 16 & $3\times 3$ & 1 \\
    Fully-connected & 256 & /\ & /\ \\
    Fully-connected & 4 & /\ & /\ \\
    \hline
  \end{tabular}
  \caption{Binary Pac-Man Network Architecture Parameters}
  \label{tab:Binary Architecture}
\end{table}

\setlength{\tabcolsep}{2pt}
\begin{table}[t]
  \centering
  \begin{tabular}{ccccc}
  \hline
    Type & Kernels & size & Stride \\
    \hline
    Convolutional & 32 & $3\times 3$ & 1 \\
    Convolutional & 32 & $3\times 3$ & 1 \\
    Convolutional & 32 & $3\times 3$ & 1 \\
    Convolutional & 32 & $3\times 3$ & 1\\
    LSTM & /\ & $32 \times 5 \times 5$ & /\\\
    Fully-connected & 256 & /\ & /\ \\
    \hline
  \end{tabular}
  \caption{Atari Games Network Architecture Parameters}
  \label{tab:Atari Architecture}
\end{table}

\subsection*{Technical Appendix B}
\label{B}

\begin{figure}[t]
    \centering
    \includegraphics[scale=0.3]{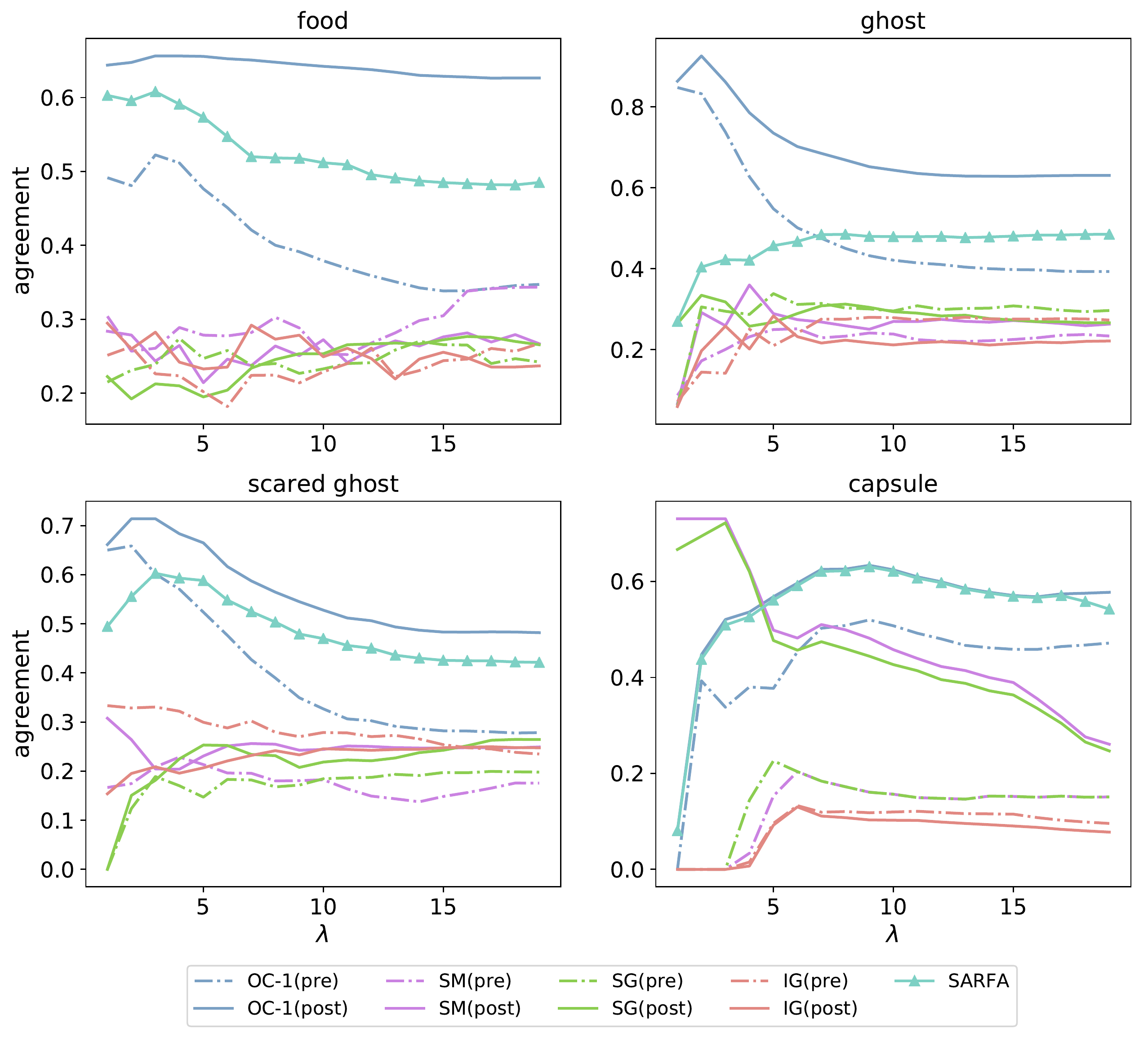}
    \caption{$\lambda$-agreement curves comparing pre-softmax and post-softmax scores for four features on Binary Pac-Man environment.}
    \label{fig: Bipacman game}
\end{figure}

\begin{figure}[t]
    \centering
    \includegraphics[scale=0.3]{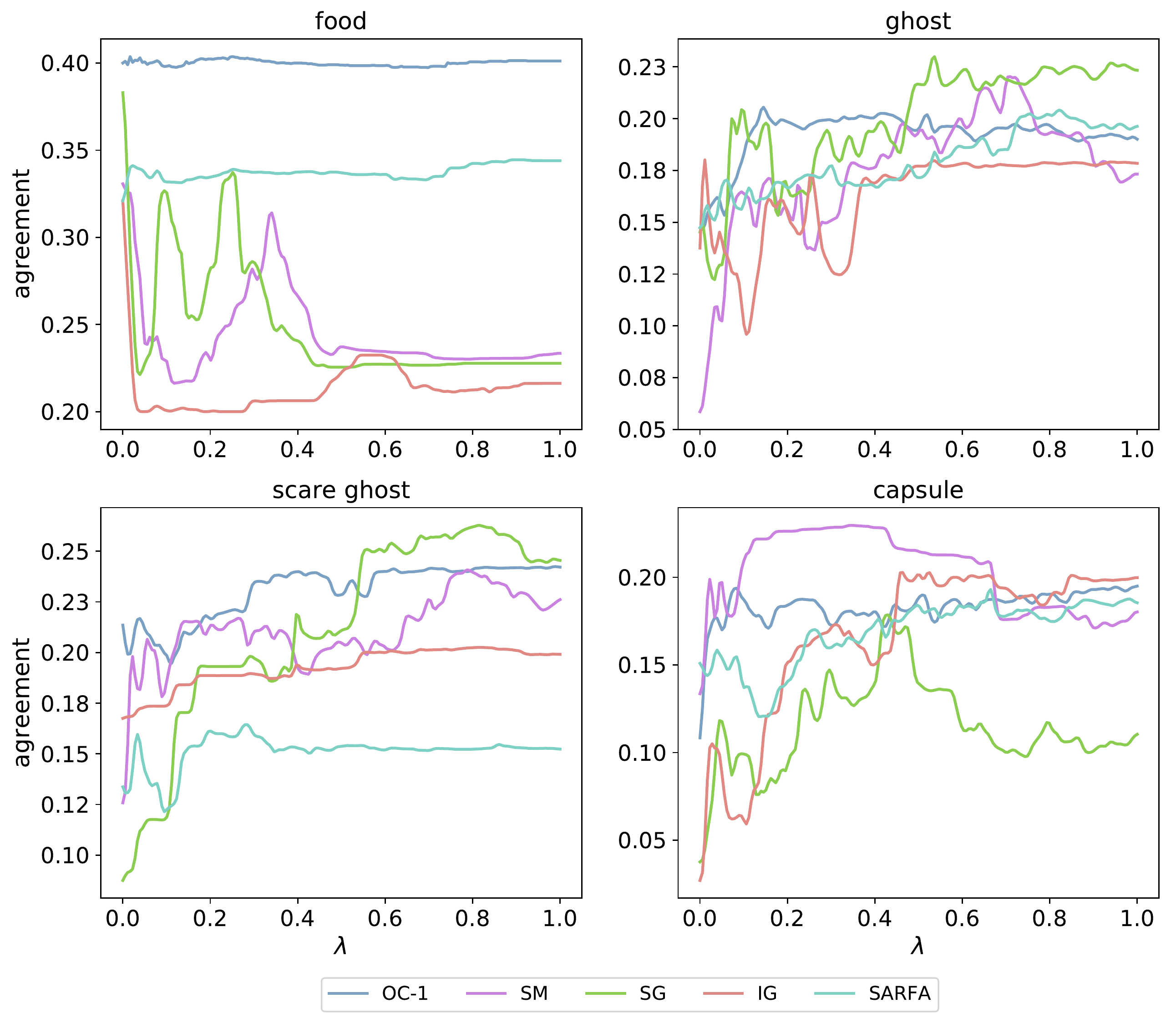}
    \caption{$\lambda$-agreement curves for four features on image Pac-Man environment.}
    \label{fig: image pacman game}
\end{figure}

In this section, we give agreement scores using $\lambda-LAR$ for all four features on binary Pac-Man (Figure \ref{fig: Bipacman game}) and image Pac-Man (Figure \ref{fig: image pacman game}). On both games, OC-1 performs the best when reconstructing actions from the food feature. When switching from binary Pac-Man to image input Pac-Man, the gradient-based attribution methods perform better than they are on the binary Pac-Man, especially for the ghost and scare ghost features since the inputs are no longer binary location matrices and they can reflect the relative importance of images to the output labels based on the gradient information calculated from the images.

\subsection*{Technical Appendix C}
\label{C}
We give the full results for Table~\ref{tab: agreement}. We repeat the same evaluation process on MsPacman, Breakout, and Space Invaders using $\lambda-LAR$ and $KAR$ on five different attribution methods. Unlike the binary Pac-Man game which takes binary matrices as inputs, DQNs for MsPacman, Breakout, and Space Invaders in Atari Environments take image input. Therefore, our treatment for these games is different. Instead of performing OC-1 on each pixel, we use the object identification~\cite{Iyer2018TransparencyAE} to locate a subset of features that are semantically meaningful to the agent. All results are evaluated over 6000 frames sampled from 20 matches. Especially, for Pac-Man game, since there are four different features, the scores are for food features in the table.

\paragraph{MsPacman.} Given the feature shape is much more complicated in MsPacman, we perform object identification by template matching with OpenCV~\cite{itseez2015opencv}. After obtaining the central locations of features and agent, we compute the $l_1$ distance between them and aggregate the number of different actions within a specific distance to calculate $\max_{\lambda, G} L(\hat{f}_\lambda(G))$. For $\lambda-LAR$, after performing a negative exponential kernel, we compute the scores for all possible actions and check whether the one with the highest score is the same as the actual action that the agent takes.

\paragraph{Space Invader.} Like MsPacman, we perform object identification by template matching with OpenCV first and use $r(\mathbf{p}, \mathbf{q}) = ||\mathbf{p}- \mathbf{q}||_2$ given that the possibility of being attacked increases with the decrease of $l_2$ distance between the agent and invaders. There is only one feature in the game which is the invader ship so for each frame we obtain the central location of different invader ships and agent then calculate $\lambda-LAR$.

\paragraph{Breakout.} Instead of select each feature and perform OC-1 or SARFA, we identify the whole \texttt{brick} in Breakout and perform OC-N (N equals the number of pixels in a \texttt{brick}). As features in Breakout are rectangular, we, therefore, use a sliding window with fixed size to select each \texttt{brick}, replace all pixels with zeros, and perform different attribution methods. We define the distance function $r$ between a brick and the agent as one plus the number of bricks between the brick of interest and the agent, given the ball will get reflected once it hits a brick. Since there is only one feature in Space Invaders (Invader ships) and Breakout (bricks), the reported scores are for these single features.

\setlength{\tabcolsep}{3pt}
\begin{table}[t]
    \centering
\begin{tabular}{ccccccc}
{Game} &OC-1 &SARFA & SM &SG&IG \\
\hline\hline
bPM & \textbf{.68}     &.65 & .27 &.27 & .29  \\
iPM  & \textbf{.32}    & .31 & .17 &.18 & .15  \\
iPM-idqn & \textbf{.31}& .27 & .17 & .16 & .13  \\
SI & .31               & .27 & .29 &\textbf{.33} & .21  \\
Br & \textbf{.33}     & .23  & .28 &.32 & .27 \\
\end{tabular}
    \caption{Evaluation of $KAR$ on different attribution methods}
    \label{tab: kernel agreement}
\end{table}

\setlength{\tabcolsep}{3pt}
\begin{table}[t]
    \centering
\begin{tabular}{ccccccc}
{Game} &OC-1 &SARFA & SM &SG&IG \\
\hline\hline
bPM & \textbf{.66} &.52  & .28 &.28 & .30   \\
iPM  & \textbf{.40}  & .34 & .33 &.38 & .32  \\
iPM-idqn &.32 & \textbf{.33} & .27 & .26 & .30  \\
SI & .34  & .35 & \textbf{.38} &.29 & .27  \\
Br & \textbf{.46} & .34  & .44 &.30 & .38 \\
\end{tabular}
    \caption{Evaluation of $\lambda-LAR$ on different attribution methods}
    \label{tab: kernel agreement}
\end{table}


\subsection*{Technical Appendix D}
\label{D}
\begin{figure}[t]
    \centering
    \includegraphics[scale=0.35]{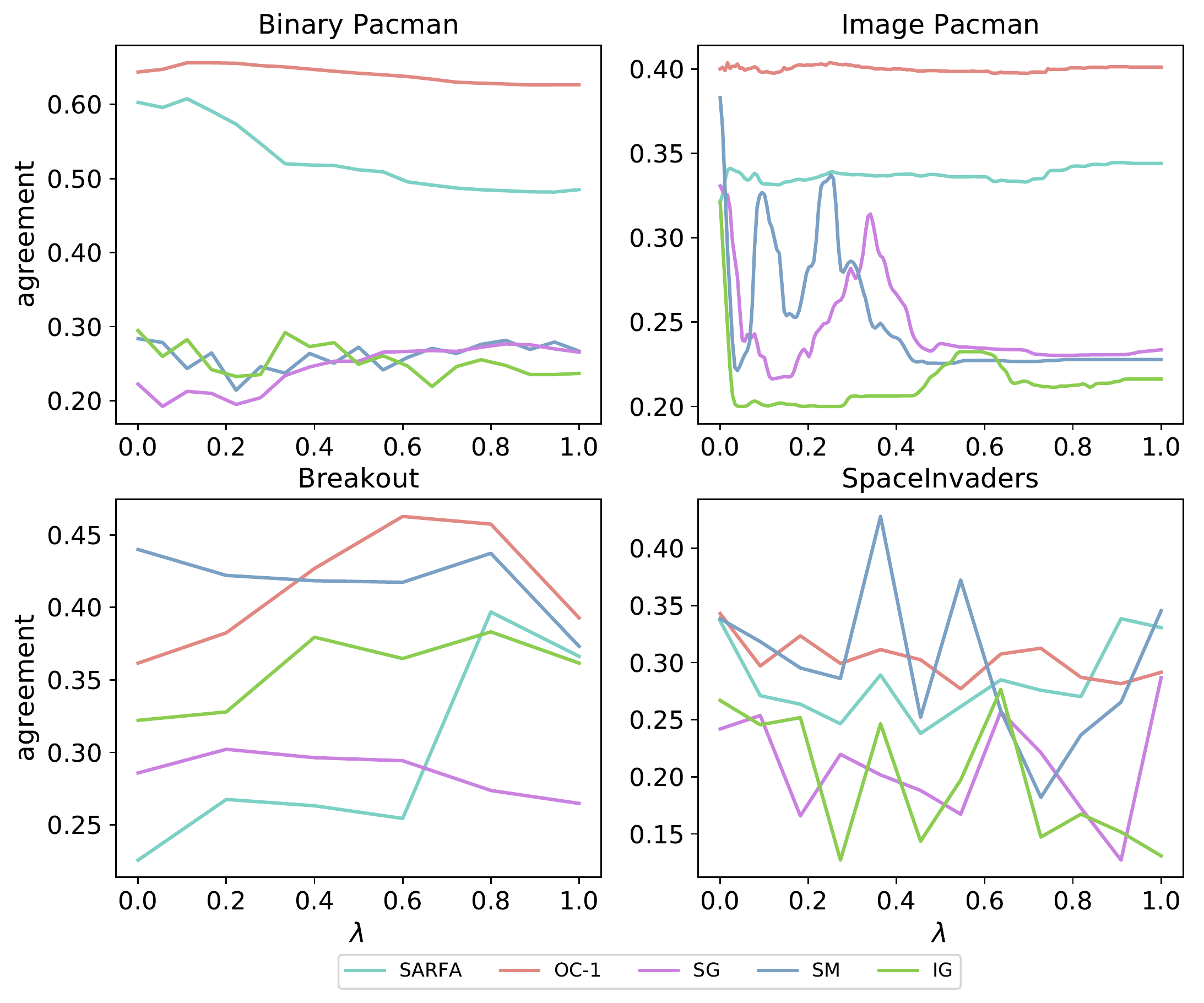}
    \caption{$\lambda$-agreement curves comparing 5 approaches on Binary Pac-Man and Atari games.}
    \label{fig: agreement compare atari}
\end{figure}

\begin{figure}[t]
    \centering
    \includegraphics[scale=0.3]{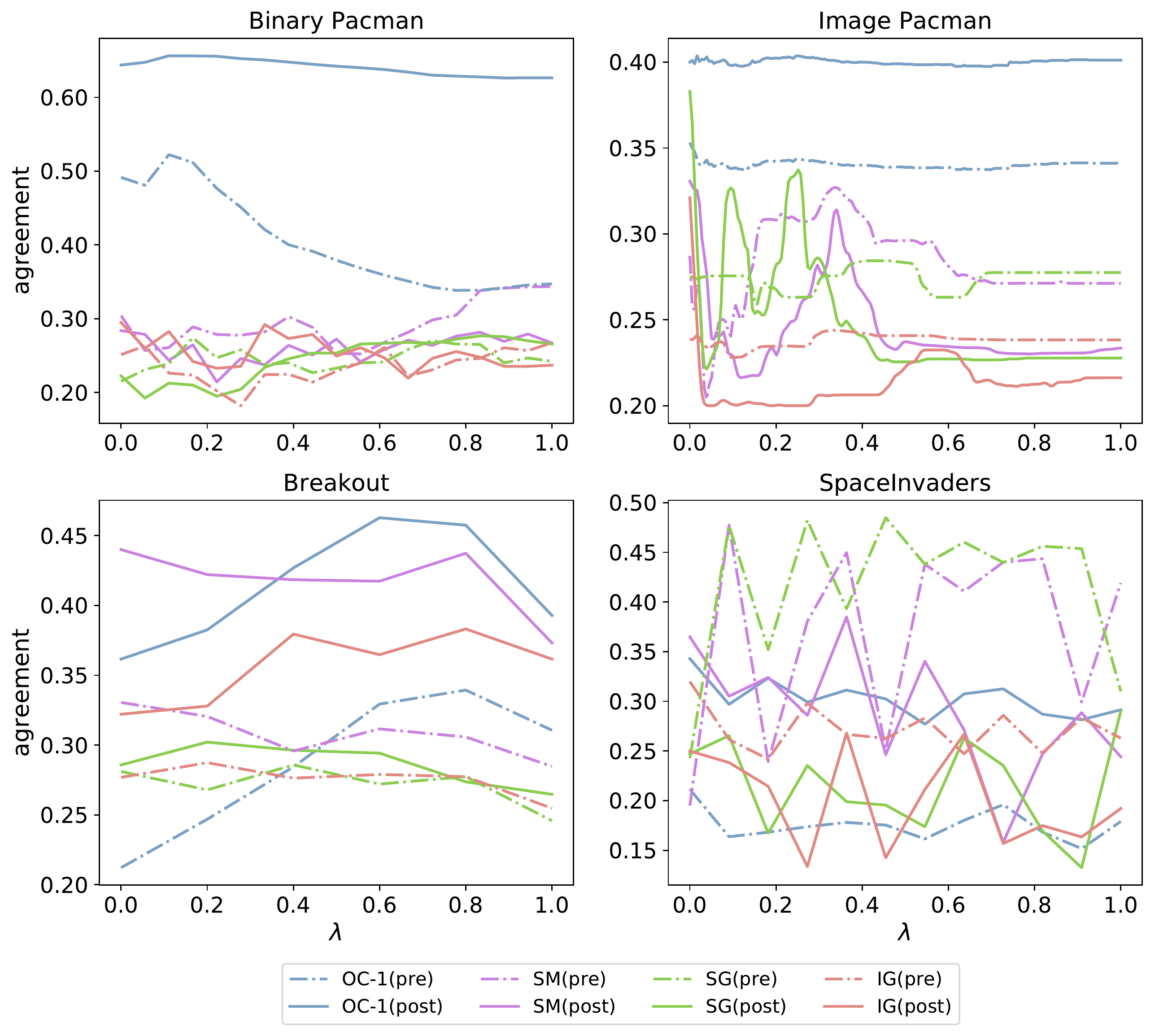}
    \caption{$\lambda$-agreement curves comparing approaches on Binary Pac-Man and Atari games. For Space Invaders, OC-1 assigns high agreement to smaller $\lambda$ which is consistent with our intuition since closer the invaders, more dangerous it is. For Breakout, it seems that OC-1, SM, and IG assign higher agreement to relatively bigger $\lambda$, indicating that a well-trained agent focuses more on the thicker part of the bricks and aims to break it out.}
    \label{fig: Atari games}
\end{figure}

In this section, we compare the performance of pre-softmax scores and post-softmax scores for SM, SG, IG, and OC-1 on binary Pac-Man, image Pac-Man, Space Invaders, and Breakout. In Figure \ref{fig: Bipacman game}, we compare two scores for four features on binary Pac-Man. For SM, SG, and IG, we take the layer before softmax and do the computation of the gradient respectively. Post-softmax score outperforms and the pre-softmax score for OC-1 is comparable for the left three gradient-based attribution approaches. In Figure \ref{fig: Atari games}, we give a comparison for food features on two Pac-Man games. For other games, as mentioned before, since there's only one feature, we use to report the scores based on that single feature. Similarly, post-softmax score outperforms the pre-softmax score for OC-1 on four games. In Figure \ref{fig: agreement compare atari}, we compare all post-softmax scores and SARFA.

\subsection*{Technical Appendix E}
\label{E}
\begin{figure}[t]
	\centering
	\begin{subfigure}{.22\textwidth}
		\centering
		\includegraphics[scale=0.3]{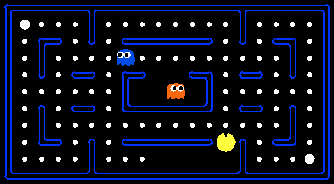}
		\caption{original input (go\_north)}
	\end{subfigure}
	\begin{subfigure}{.22\textwidth}
		\centering
		\includegraphics[scale=0.17]{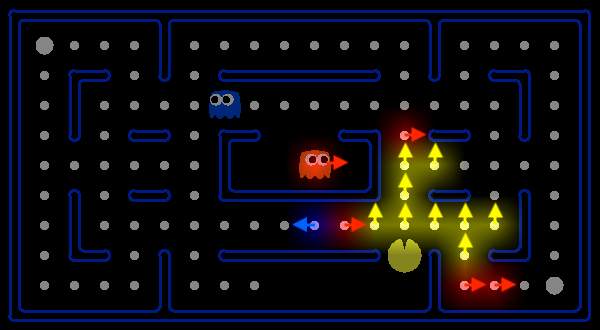}
		\caption{OC-1}
	\end{subfigure}

	\begin{subfigure}{.22\textwidth}
		\centering
		\includegraphics[scale=0.17]{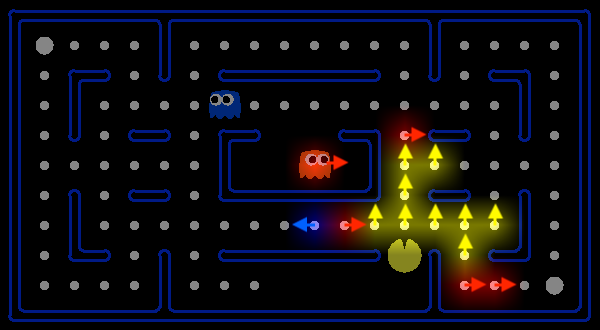}
		\caption{SARFA}
	\end{subfigure}
	\begin{subfigure}{.22\textwidth}
		\centering
		\includegraphics[scale=0.17]{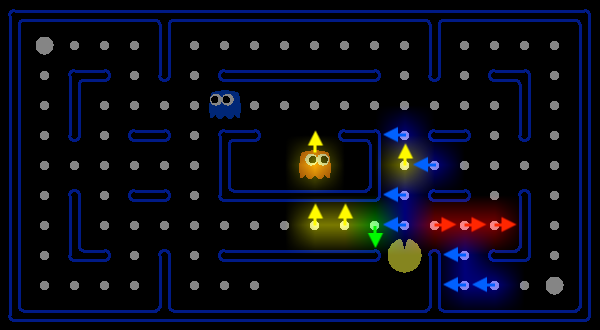}
		\caption{IG}
	\end{subfigure}

	\begin{subfigure}{.22\textwidth}
		\centering
		\includegraphics[scale=0.17]{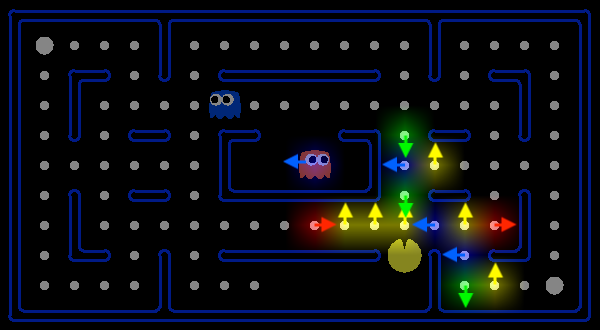}
		\caption{SG}
	\end{subfigure}
	\begin{subfigure}{.22\textwidth}
		\centering
		\includegraphics[scale=0.17]{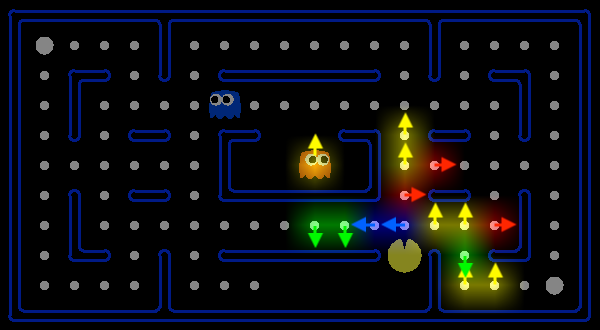}
		\caption{SM}
	\end{subfigure}
	\caption{}
	\label{fig: behavior-level1}
\end{figure}

\begin{figure}[t]
	\centering
	\begin{subfigure}{.22\textwidth}
		\centering
		\includegraphics[scale=0.17]{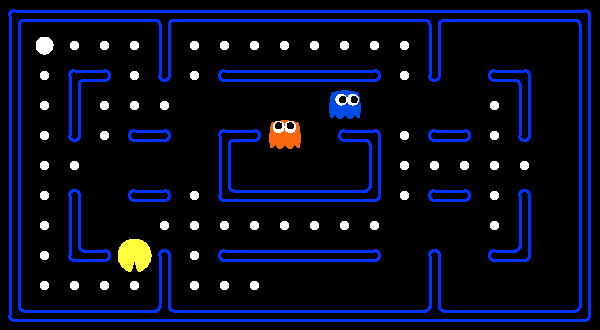}
		\caption{original input (go\_south)}
	\end{subfigure}
	\begin{subfigure}{.22\textwidth}
		\centering
		\includegraphics[scale=0.17]{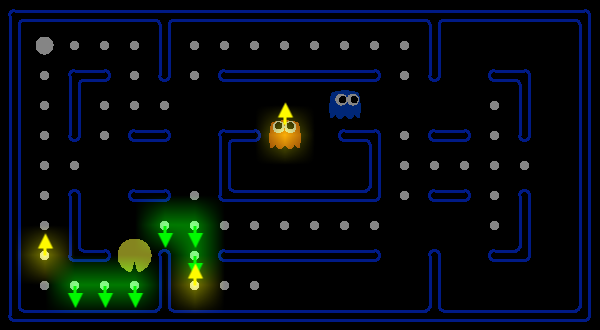}
		\caption{OC-1}
	\end{subfigure}

	\begin{subfigure}{.22\textwidth}
		\centering
		\includegraphics[scale=0.17]{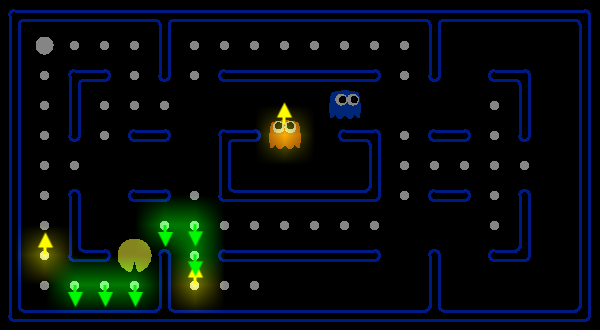}
		\caption{SARFA}
	\end{subfigure}
	\begin{subfigure}{.22\textwidth}
		\centering
		\includegraphics[scale=0.17]{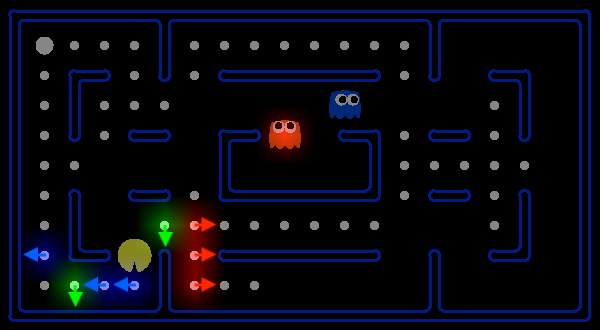}
		\caption{IG}
	\end{subfigure}

	\begin{subfigure}{.22\textwidth}
		\centering
		\includegraphics[scale=0.17]{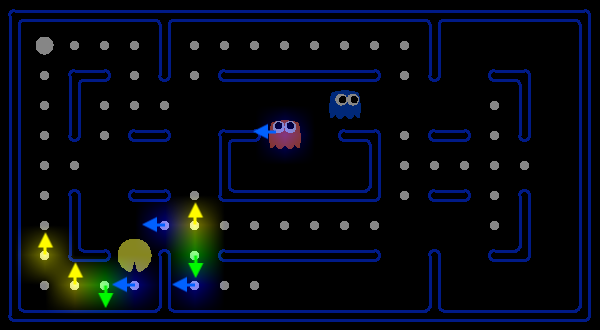}
		\caption{SG}
	\end{subfigure}
	\begin{subfigure}{.22\textwidth}
		\centering
		\includegraphics[scale=0.17]{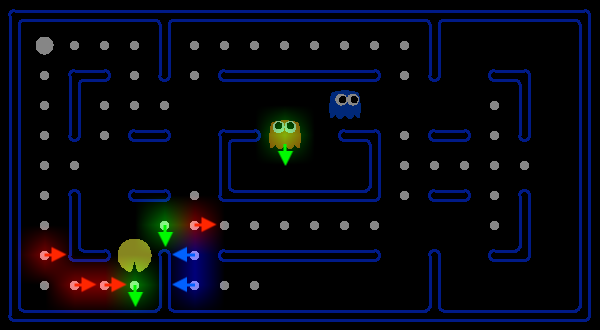}
		\caption{SM}
	\end{subfigure}
	\caption{}
	\label{fig: behavior-level2}
\end{figure}

\begin{figure}[t]
	\centering
	\begin{subfigure}{.22\textwidth}
		\centering
		\includegraphics[scale=0.17]{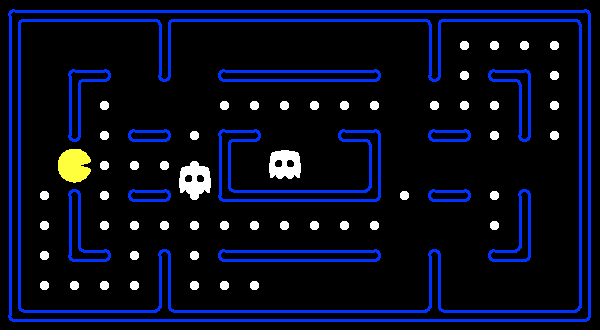}
		\caption{original input (go\_east)}
	\end{subfigure}
	\begin{subfigure}{.22\textwidth}
		\centering
		\includegraphics[scale=0.17]{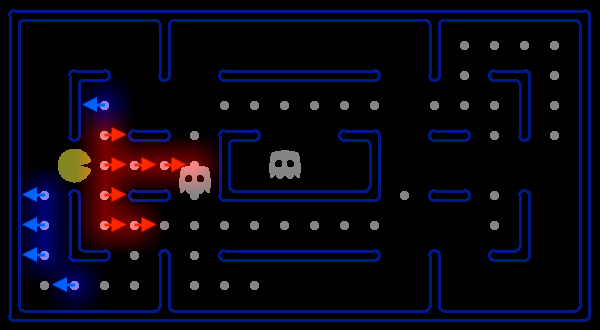}
		\caption{OC-1}
	\end{subfigure}

	\begin{subfigure}{.22\textwidth}
		\centering
		\includegraphics[scale=0.17]{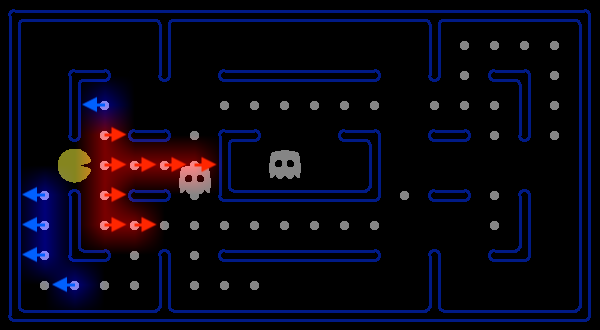}
		\caption{SARFA}
	\end{subfigure}
	\begin{subfigure}{.22\textwidth}
		\centering
		\includegraphics[scale=0.17]{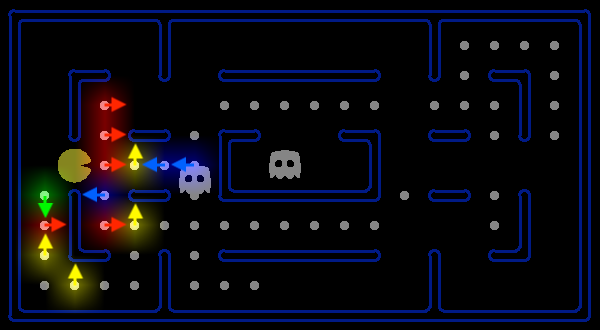}
		\caption{IG}
	\end{subfigure}

	\begin{subfigure}{.22\textwidth}
		\centering
		\includegraphics[scale=0.17]{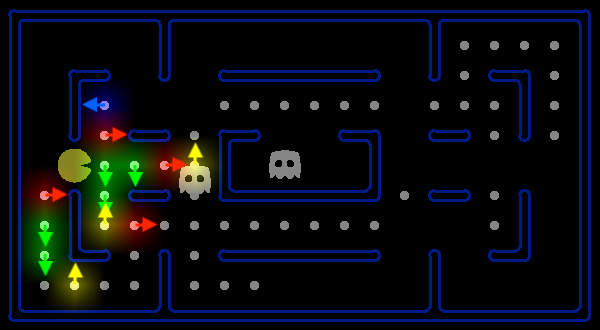}
		\caption{SG}
	\end{subfigure}
	\begin{subfigure}{.22\textwidth}
		\centering
		\includegraphics[scale=0.17]{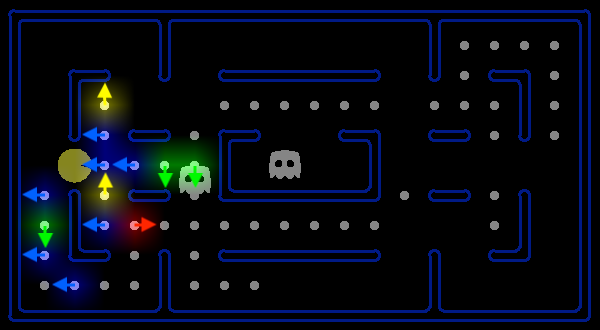}
		\caption{SM}
	\end{subfigure}
	\caption{}
	\label{fig: behavior-level3}
\end{figure}

\begin{figure}[!t]
	\centering
	\begin{subfigure}{.22\textwidth}
		\centering
		\includegraphics[scale=0.17]{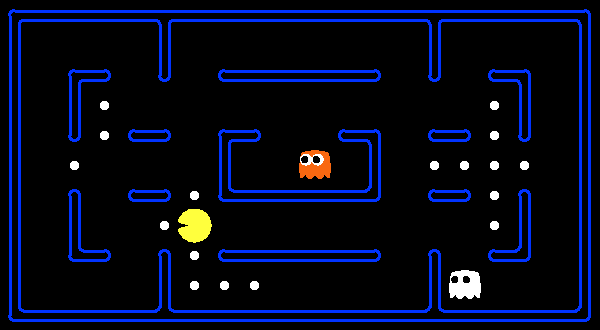}
		\caption{original input (go\_south)}
	\end{subfigure}
	\begin{subfigure}{.22\textwidth}
		\centering
		\includegraphics[scale=0.17]{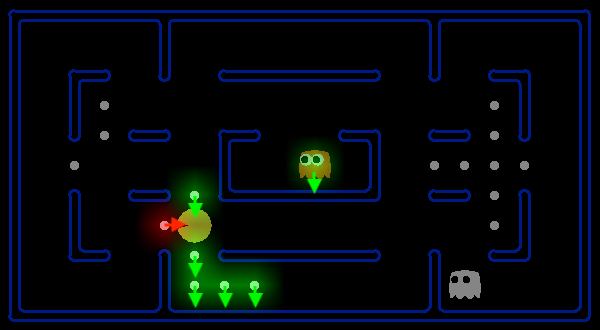}
		\caption{OC-1}
	\end{subfigure}

	\begin{subfigure}{.22\textwidth}
		\centering
		\includegraphics[scale=0.17]{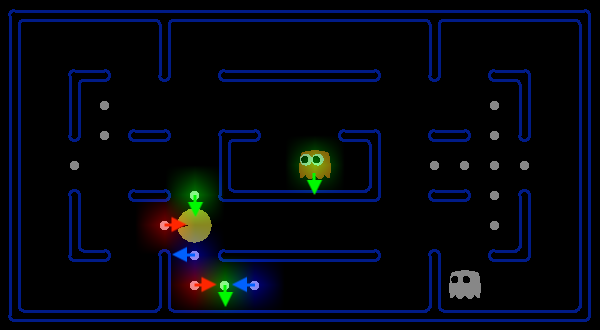}
		\caption{SARFA}
	\end{subfigure}
	\begin{subfigure}{.22\textwidth}
		\centering
		\includegraphics[scale=0.17]{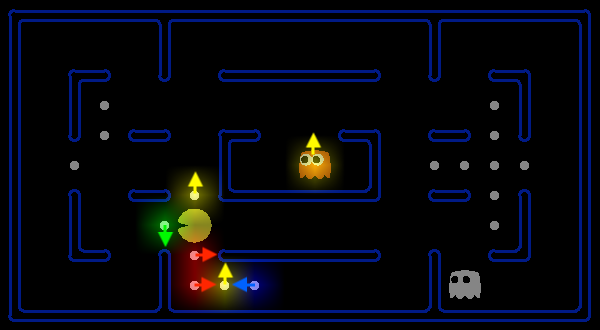}
		\caption{IG}
	\end{subfigure}

	\begin{subfigure}{.22\textwidth}
		\centering
		\includegraphics[scale=0.17]{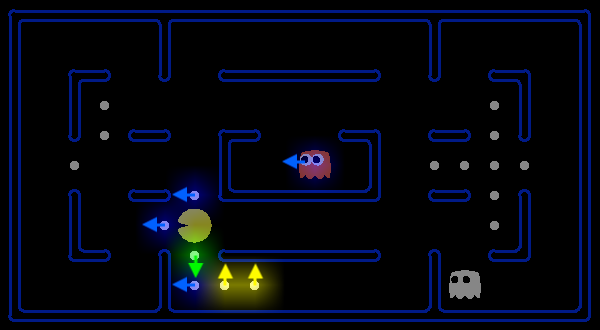}
		\caption{SG}
	\end{subfigure}
	\begin{subfigure}{.22\textwidth}
		\centering
		\includegraphics[scale=0.17]{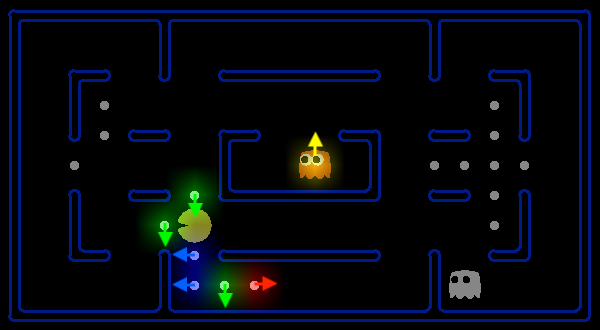}
		\caption{SM}
	\end{subfigure}
	\caption{}
	\label{fig: behavior-level4}
\end{figure}

\begin{figure}[!h]
	\centering
	\begin{subfigure}{.22\textwidth}
		\centering
		\includegraphics[scale=0.17]{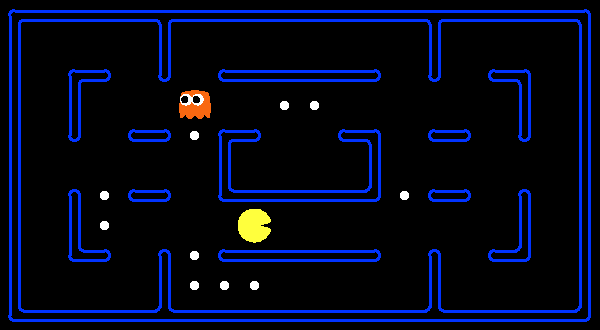}
		\caption{original input (go\_west)}
	\end{subfigure}
	\begin{subfigure}{.22\textwidth}
		\centering
		\includegraphics[scale=0.17]{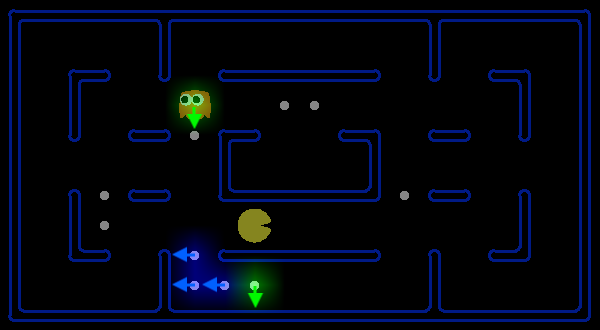}
		\caption{OC-1}
	\end{subfigure}

	\begin{subfigure}{.22\textwidth}
		\centering
		\includegraphics[scale=0.17]{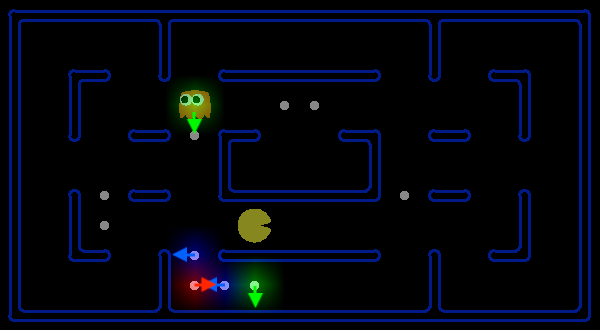}
		\caption{SARFA}
	\end{subfigure}
	\begin{subfigure}{.22\textwidth}
		\centering
		\includegraphics[scale=0.17]{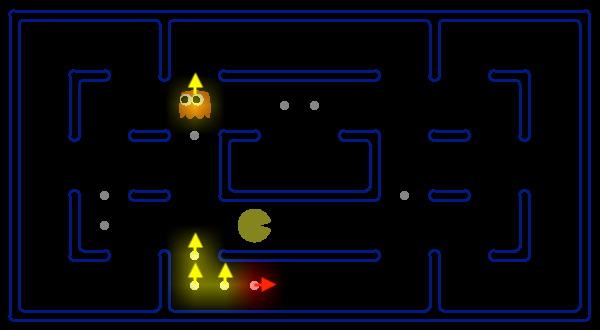}
		\caption{IG}
	\end{subfigure}

	\begin{subfigure}{.22\textwidth}
		\centering
		\includegraphics[scale=0.17]{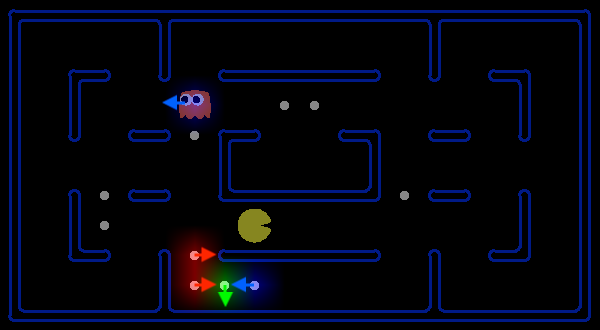}
		\caption{SG}
	\end{subfigure}
	\begin{subfigure}{.22\textwidth}
		\centering
		\includegraphics[scale=0.17]{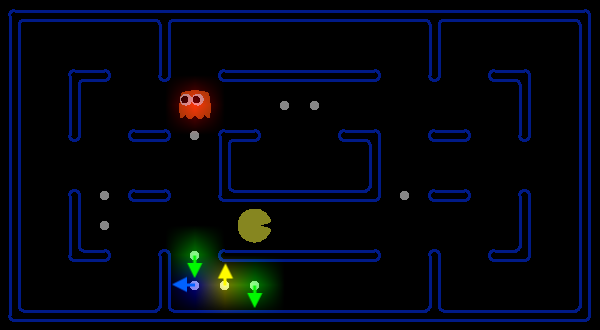}
		\caption{SM}
	\end{subfigure}
	\caption{}
	\label{fig: behavior-level5}
\end{figure}

\begin{figure}[!t]
	\centering
	\begin{subfigure}{.22\textwidth}
		\centering
		\includegraphics[scale=0.17]{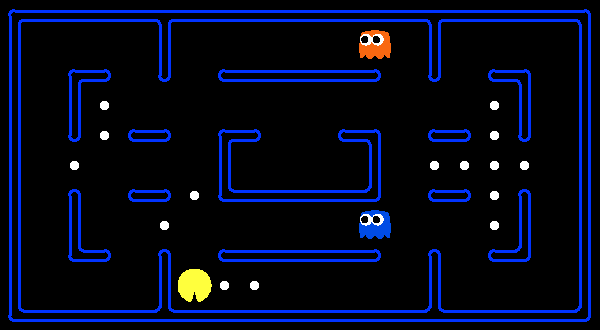}
		\caption{original input (go\_east)}
	\end{subfigure}
	\begin{subfigure}{.22\textwidth}
		\centering
		\includegraphics[scale=0.17]{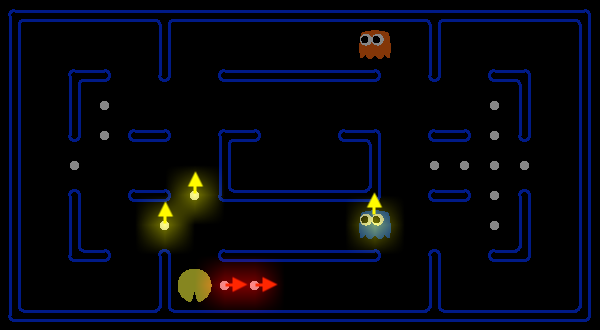}
		\caption{OC-1}
	\end{subfigure}

	\begin{subfigure}{.22\textwidth}
		\centering
		\includegraphics[scale=0.17]{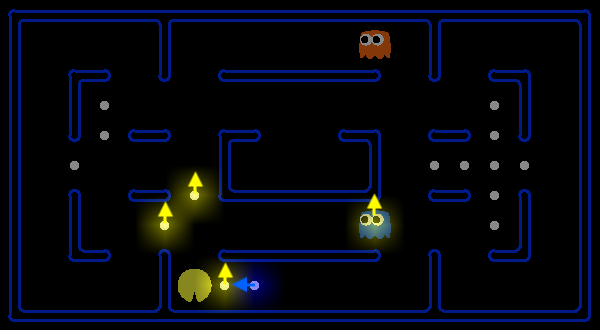}
		\caption{SARFA}
	\end{subfigure}
	\begin{subfigure}{.22\textwidth}
		\centering
		\includegraphics[scale=0.17]{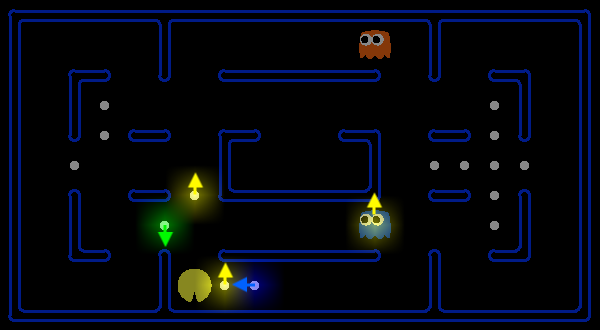}
		\caption{IG}
	\end{subfigure}

	\begin{subfigure}{.22\textwidth}
		\centering
		\includegraphics[scale=0.17]{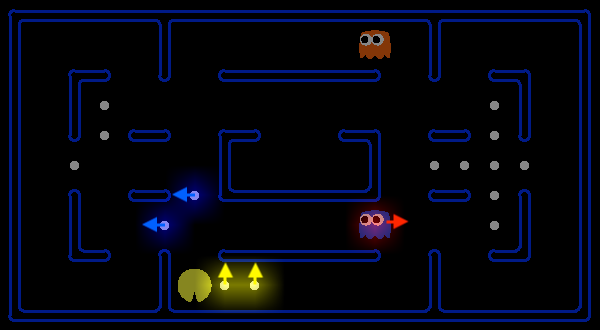}
		\caption{SG}
	\end{subfigure}
	\begin{subfigure}{.22\textwidth}
		\centering
		\includegraphics[scale=0.17]{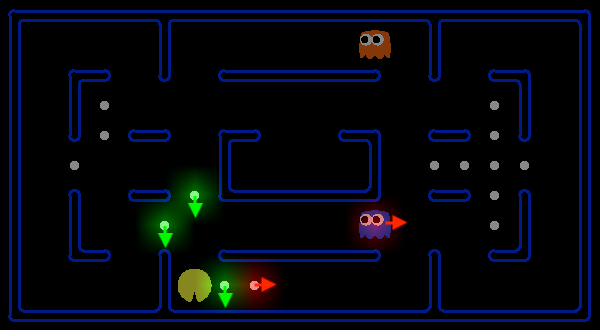}
		\caption{SM}
	\end{subfigure}
	\caption{}
	\label{fig: behavior-level6}
\end{figure}

In this section, we provide more visualizations of the behavior-level attributions on binary Pac-Man game (Fig.~\ref{fig: behavior-level1} to \ref{fig: behavior-level6}). We find SARFA and OC-1 with post-softmax scores tend to produce similar results when there are many \texttt{food} in the environment (Fig.~\ref{fig: behavior-level1} to \ref{fig: behavior-level3}). But when the game is close the termination (Fig.~\ref{fig: behavior-level4} to ~\ref{fig: behavior-level6}), where \texttt{food} becomes less, SARFA starts to assign different directions to nearby features other than the actual action. One possible reason is that when \text{food} becomes less, $K$ in SARFA becomes dominant. As a result, SARFA tends to indicate the importance of this feature towards all other actions when it is perturbed instead of the importance towards the action of interest. But why and how the number of \text{food} influences the behavior of $K$ to remain unknown to us. We consider this result as discovering one of the limitations of SARFA.

\end{document}